\documentclass[journal,10pt]{IEEEtran}
\usepackage{epsfig,latexsym}
\usepackage{float}
\usepackage{indentfirst}
\usepackage{amsmath}
\usepackage{amssymb}
\usepackage{times}
\usepackage{psfrag}
\usepackage{cite}
\usepackage{subfigure}
\usepackage{textcomp}
\usepackage{multirow}
\usepackage{multicol}
\usepackage{stfloats}
\usepackage{url}
\usepackage{subfigure}
\usepackage{multirow}
\usepackage{booktabs}
\usepackage{threeparttable}

\usepackage{fancyhdr}
\begin{document}




\title{Predicting Video Saliency with Object-to-Motion CNN and Two-layer Convolutional LSTM}

\author{Lai~Jiang,  ~\IEEEmembership{Student Member,~IEEE,} Mai~Xu, ~\IEEEmembership{Senior Member,~IEEE,} and Zulin~Wang, ~\IEEEmembership{Member,~IEEE}
\thanks{L. Jiang, M. Xu and Z. Wang are with the School of Electronic and Information Engineering, Beihang University, Beijing, 100191 China (e-mail: Maixu@buaa.edu.cn;  jianglai.china@buaa.edu.cn; wzulin@buaa.edu.cn). This work was supported by the China 973 Program under Grant 2013CB329006, the National Nature Science Foundation of China projects under Grants 61573037 and 61202139, and the Fok Ying Tung Education Foundation under Grant 151061. This work was also supported by the MSRA visiting young faculty program. Mai Xu is the corresponding author of this paper.}}

\maketitle

\begin{abstract}
Over the past few years, deep neural networks (DNNs) have exhibited great success in predicting the saliency of images. However, there are few works that apply DNNs to predict the saliency of generic videos. In this paper, we propose a novel DNN-based video saliency prediction method. Specifically, we establish a large-scale eye-tracking database of videos (LEDOV), which provides sufficient data to train the DNN models for predicting video saliency. Through the statistical analysis of our LEDOV database, we find that human attention is normally attracted by objects, particularly moving objects or the moving parts of objects. Accordingly, we propose an object-to-motion convolutional neural network (OM-CNN) to learn spatio-temporal features for predicting the intra-frame saliency via exploring the information of both objectness and object motion. We further find from our database that there exists a temporal correlation of human attention with a smooth saliency transition across video frames. Therefore, we develop a two-layer convolutional long short-term memory (2C-LSTM) network in our DNN-based method, using the extracted features of OM-CNN as the input. Consequently, the inter-frame saliency maps of videos can be generated, which consider the transition of attention across video frames. Finally, the experimental results show that our method advances the state-of-the-art in video saliency prediction.
%
%
\end{abstract}
%

\section{Introduction}

A foveated mechanism \cite{matin1974saccadic} in the Human Visual System (HVS) indicates that only small fovea region captures most visual attention at high resolution, while other peripheral regions receive little attention at low resolution. To predict human attention, saliency detection has been widely studied in recent years, with multiple applications \cite{borji2013state} in object recognition, object segmentation, action recognition, image caption, image/video compression, etc. In this paper, we focus on predicting video saliency at pixel level, which models attention on each video frame.

In the early time,the traditional methods of video saliency prediction mainly follow the integration theory \cite{itti2004realistic,ren2013regularized,Nguyen2013cameramotion,Zhong2013dynamicconsis,lee2014video}, i.e., saliency of video frames can be detected by two steps: (1) Extract spatial and temporal features from videos for obtaining conspicuous maps; (2) Conduct a fusion strategy to combine conspicuous maps of different feature channels together for generating saliency maps. Benefitting from the state-of-the-art image saliency prediction, a great deal of spatial features have been incorporated to predict video saliency \cite{judd2009learning,cheng2015global,zhang2016exploiting}. Additionally, some works focused on designing temporal features for video saliency prediction, mainly in three aspects: motion based features \cite{harel2006graph,Zhong2013dynamicconsis,zhou2014time}, temporal contrast features \cite{itti2004realistic,zhang2009sunday,ren2013regularized} and compressed domain features \cite{Khatoonabadi2015CVPR,xu2017learning}. For fusing spatial and temporal features, many machine learning algorithms were utilized, such as Support Vector Machine (SVM) \cite{judd2009learning,lee2014video,xu2017learning}, probabilistic model \cite{itti2009bayesian,zhang2009sunday,rudoy2013iccv} and phase spectrum analysis \cite{guo2010novel,zhang2015unsupervised}.

Different from the integration theory, deep natural networks (DNN) based methods have been recently proposed to learn human attention in an end-to-end manner, significantly boosting the accuracy of image saliency prediction \cite{KummererTB14,kruthiventi2015deepfix,huang2015salicon,pan2016shallow,li2016deepsaliency,wang2016RCNN,Pan_2017_SalGAN}.
However, only a few works managed to apply DNN in video saliency prediction \cite{bak2016two,bazzani2016recurrent,wang2017deep}. Specifically, Cagdas \emph{et al.} \cite{bak2016two} applied a two-stream CNN structure taking both RGB frames and motion maps as the inputs, for video saliency prediction. Bazzani \emph{et al.} \cite{bazzani2016recurrent} leveraged a deep Convolutional 3D (C3D) network to learn the representations of human attention on 16 consecutive frames, and then a Long Short-Term Memory (LSTM) network connected with a mixture density network was learned to generate saliency maps in Gaussian mixture distribution. However, the above DNN based methods\footnote{None of the above methods makes the code available online. In contrast, our code is accessible in https://github.com/remega/OMCNN\_2CLSTM.} for video saliency prediction is still in infancy due to the following drawbacks: (1) Insufficient eye-tracking data for training DNN; (2) Lack of sophisticated network architecture simultaneously learning to combine information of object and motion; (3) Neglect of dynamic pixel-wise transition of video saliency across video frames.

To avoid the above drawbacks, this paper proposes a new DNN based method to predict video saliency with spatio-temporal representation and dynamic saliency modeling, benefitting from the analysis of our eye-tracking database. Specifically, we first establish a large-scale video database, which contains eye-tracking data of 32 subjects on viewing 538 diverse-content videos. Through analysis on our database, we find that human attention is normally attracted by objects in a video, especially by the moving objects or moving parts in the objects. In light of this finding, a novel Object-to-Motion Convolutional Neural Network (OM-CNN) is constructed to learn spatio-temporal features for video saliency prediction, considering both objectness and object motion information. For dynamic saliency modeling, a Two-layer Convolutional LSTM (2C-LSTM) network is developed to predict the pixel-wise transition of video saliency across frames, with input of spatio-temporal features from OM-CNN. Different from the conventional LSTM network, this structure is capable of keeping spatial information through the convolutional connections.

To summarize, the main contributions of our work are listed in the following:
\begin{itemize}
\item We establish an eye tracking database consisting of 538 videos in diverse content, with the thorough analysis and findings on our database.
\item We propose the novel OM-CNN structure to predict saliency of intra-frame, which integrates both objectness and object motion in a uniform deep structure.
\item We develop the 2C-LSTM network with Bayesian dropout to learn the saliency transition across inter-frame at pixel-wise level.
\end{itemize}

The rest of this paper is organized as follows. In Section \ref{related}, we briefly review the related works and eye-tracking databases for video saliency prediction. In Section \ref{reviewdatabase}, we establish and analyze our large-scale eye-tracking database. According to the findings on our database, we propose a DNN for video saliency prediction in Section \ref{method}, including both OM-CNN and 2C-LSTM. Section \ref{experiment} shows the experiment results to validate the performance of our method. Section \ref{conclusion} concludes this paper.

\section{Related work}\label{related}
In this section, we briefly review the recent works and eye-tracking databases for video saliency prediction.

\subsection{Video saliency prediction}\label{videosaliency}

Most traditional methods for video saliency prediction  \cite{itti2004realistic,ren2013regularized,Nguyen2013cameramotion,Zhong2013dynamicconsis,lee2014video} rely on the integration theory consisting of two main steps: feature extraction and feature fusion. In the task of image saliency prediction, many effective spatial features succeed in predicting human attention with  either top-down \cite{judd2009learning,goferman2012context} or bottom-up \cite{itti1998model,cheng2015global} strategy. However, video saliency prediction is more challenging, because temporal features also play an important role in drawing human attention. To achieve this, motion based features \cite{harel2006graph,Zhong2013dynamicconsis,zhou2014time}, temporal difference \cite{itti2004realistic,zhang2009sunday,ren2013regularized} and compressed domain methods \cite{Fang2014tcsvt,xu2017learning} are widely used in the existing works of video saliency prediction. Taking motion as additional temporal features, Zhong \emph{et al.} \cite{Zhong2013dynamicconsis} proposed to predict video saliency using modified optical flow with restriction of dynamic consistence. Similarly, Zhou \emph{et al.} \cite{zhou2014time} extended motion feature by computing center motion, foreground motion, velocity motion and acceleration motion in their saliency prediction method. In addition to motion, other methods \cite{itti2004realistic,zhang2009sunday,ren2013regularized} make use of the temporal changes in videos for saliency prediction, through figuring out the contrast between successive frames. For example, Ren \emph{et al.} \cite{ren2013regularized} proposed to estimate the temporal difference of each patch by finding the minimal reconstruction error of sparse representation over the co-located patches of neighboring frames. Similarly, in \cite{zhang2009sunday}, the temporal difference is obtained via adding pre-designed exponential filters on spatial features of successive frames. Taking advantage from sophisticated video coding standards, the compressed domain features are also explored as spatio-temporal features for video saliency prediction \cite{Fang2014tcsvt,xu2017learning}.

In addition to feature extraction, many works focus on the fusion strategy to generate video saliency maps. Specifically, a set of probability models were constructed to calculate the posterior/prior beliefs \cite{itti2009bayesian}, joint probability distribution of features \cite{zhang2009sunday} and candidate transition probability \cite{rudoy2013iccv}, in predicting video saliency. Similarly, Li \emph{et al.} \cite{li2010probabilistic} developed a probabilistic multi-task learning method to incorporate the task-related prior in video saliency prediction. Besides, other machine learning algorithms, such as SVM and neutral network, were also applied for linearly \cite{Nguyen2013cameramotion} or non-linearly \cite{lee2014video} combining the saliency related features. Other advanced methods \cite{guo2010novel,zhang2015unsupervised} apply phase spectrum analysis in the fusion model to bridge the gap between features and video saliency. For instance , Guo \emph{et al.} \cite{guo2010novel} applied Phase spectrum of Quaternion Fourier Transform (PQFT) on four feature channels (two color channels, one intensity channel, and one motion channel) to predict video saliency.

Most recently, DNN has succeeded in many computer vision tasks, such as image classification \cite{simonyan2014very}, action recognition \cite{du2015hierarchical} and object detection \cite{Redmon2015}. In the field of saliency prediction, DNN has also been successfully incorporated to automatically learn spatial features for predicting saliency of images \cite{kruthiventi2015deepfix,huang2015salicon,pan2016shallow,li2016deepsaliency,wang2016RCNN,Pan_2017_SalGAN,cornia2016deep}. Specifically, as one of the pioneering works, Deepfix \cite{kruthiventi2015deepfix} proposed a DNN based structure on VGG-16 \cite{simonyan2014very} and inception module \cite{szegedy2015going} to learn multi-scales semantic representation for saliency prediction. In Deppfix, a dilated convolutional structure was developed to extend receptive field, and then a location biased convolutional layer was proposed to learn the centre-bias pattern for saliency prediction. Similarly, SALICON \cite{huang2015salicon} was also proposed to fine tune the existing object recognition DNNs, and developed an efficient loss function for training the DNN model in saliency prediction. Later, some advanced DNN methods \cite{pan2016shallow,li2016deepsaliency,wang2016RCNN,cornia2016deep} were proposed to improve the performance of image saliency prediction.

However, only a few works manage to apply DNN in video saliency prediction \cite{Chaabouni2016deepsaliency,bak2016two,bazzani2016recurrent,wang2017deep,Liu_2017_CVPR}. In these DNNs, the dynamic characteristics were explored in two ways: adding temporal information in CNN structures \cite{Chaabouni2016deepsaliency,bak2016two,wang2017deep} or developing dynamic structure with LSTM \cite{bazzani2016recurrent,Liu_2017_CVPR}. For adding temporal information, a four-layer CNN in \cite{Chaabouni2016deepsaliency} and a two-stream CNN in \cite{bak2016two} were trained, respectively, with both RGB frames and motion maps as the inputs. Similarly, in \cite{wang2017deep}, the pair of video frames concatenated with a static saliency map (generated by the static CNN) are input to the dynamic CNN for video saliency prediction, allowing CNN to generalize more temporal features through the representation learning of DNN. Instead, we find that human attention is more likely to be attracted by the moving objects or moving parts of the objects. As such, to explore the semantic temporal features for video saliency prediction, a motion subnet in our OM-CNN is trained under the guidance of the objectness subnet.


For developing the dynamic structure, Bazzani \emph{et al.} \cite{bazzani2016recurrent} and Liu \emph{et al.} \cite{Liu_2017_CVPR} applied the LSTM networks to predict human attention, relying on both short- and long-term memory. However, the fully connections in LSTM limit dimensions of both input and output, unable to obtain end-to-end saliency map. As such, the strong prior knowledge need to be assumed for the distribution of saliency. To be more specific, in \cite{bazzani2016recurrent}, the human attention is assumed to distribute as Gaussian Mixture Model (GMM), then the LSTM is constructed to learn parameters of GMM. Similarly, \cite{Liu_2017_CVPR} focuses on predicting the saliency of conference videos and assume that the saliency in each face is a Gaussian distribution. In \cite{Liu_2017_CVPR}, the face saliency transition  across video frames is learned by LSTM, and the final saliency map is generated via combining the saliency of all faces in video. In our work, we first explore 2C-LSTM with Bayesian dropout, to directly predict saliency maps in an end-to-end manner. This allows learning the more complex distribution of human attention, rather than pre-assumed distribution of saliency.

\subsection{Database}\label{reviewdatabase}

The eye-tracking databases of videos collect the fixations of subjects on each video frame, which can be used as the ground truth for video saliency prediction. The existing eye-tracking databases benefit from the mature eye-tracking technology. In particular, an eye tracker is used to obtain the fixations of subjects on videos, by tracking the pupil and corneal reflections \cite{Holmqvist2011eyetracking}. The pupil locations are then mapped to the real-world stimuli, i.e., video frames, through a pre-defined calibration matrix. As such, fixations can be located in each video frame, indicating where people pay attention.

Now, we review the existing video eye-tracking databases. Table \ref{DatabaseDetail} summarizes the basic properties of these databases. To the best of our knowledge, CRCNS \cite{itti2004crcns}, SFU \cite{Hadizadeh2012SFU}, DIEM \cite{Mital2011DIEM} and Hollywood \cite{Mathe2015pami} are the most popular databases, widely used in the most of recent video saliency prediction works \cite{fang2014tip,Fang2014tcsvt,Khatoonabadi2015CVPR,rudoy2013iccv,Nguyen2013cameramotion,Zhong2013dynamicconsis,zhang2015unsupervised,Chaabouni2016deepsaliency,mauthner2015encoding}. In the following, they are reviewed in more details.

\textbf{CRCNS} \cite{itti2004crcns} is one of the earlist video eye-tracking databases established by Itti \emph{et al.} in 2004. It is still used as a benchmark in the recent video saliency prediction works, such as \cite{fang2014tip}. CRCNS contains 50 videos mainly including outdoor scenes, TV shows and video games. The length of each video ranges from 5.5 to 93.9 seconds, and the frame rate of all videos is 30 frames per second (fps). For each video, 4 to 6 subjects were asked to look at the main actors or actions. Afterward, they were required to depict the main content of video. Thus, CRNS is a task-driven eye-tracking database for videos. Later, a new database \cite{Carmi2006crcns} was established, by manually cutting all 50 videos of CRCNS into 523 ``clippets'' with 1-3 second duration, according to the abrupt cinematic cuts. Another 8 subjects were recruited to view these video clippets, with their eye-tracking data recorded in \cite{Carmi2006crcns}.

\textbf{SFU} \cite{Hadizadeh2012SFU} is a public video database containing eye-tracking data of 12 uncompressed YUV videos, which are frequently used as the standard test set for video compression and processing algorithms. Each video is in the CIF resolution ($352\times288$), and is with  3-10 seconds at a frame rate of 30 fps. All eye-tracking data were collected, when 15 non-expert subjects were free viewing all 12 videos twice.

\textbf{DIEM} \cite{Mital2011DIEM} is another widely used database, designed to evaluate the contributions of different visual features on gaze clustering. DIEM comprises 84 videos sourced from publicly accessible videos including advertisement, game trailer, movie trailer and news clip. Most of these videos have frequent cinematic cuts. Each video lasts for 27-217 seconds at 30 fps. The free-viewing fixations of around 50 subjects were tracked for each video.

\textbf{Hollywood} \cite{Mathe2015pami} is a large-scale eye-tracking database for video saliency prediction, which contains all videos from two action recognition databases: Hollywood-2 \cite{Marzakek2006action} and UCF sports \cite{rodriguez2010spatio}. All of 1707 videos in Hollywood-2 were selected from 69 movies, according to 12 action classes, such as answering phone, eating and shaking hands. UCF sports is another action database including 150 videos with 9 sport action classes. The human fixations of 19 subjects were captured under 3 conditions: free viewing (3 subjects), action recognition task (12 subjects), and context recognition task (4 subjects). Although the video number of Hollywood is large, its video content is not diverse, constrained by human actions. Besides, it mainly focuses on task-driven viewing mode, rather than free viewing.

\begin{table}[]
  \tiny
  \centering
  \caption{\footnotesize The basic properties of the existing eye-tracking databases.}\label{DatabaseDetail}
    \begin{tabular}{ccccccc}
    \toprule
    Database & Year & Videos & Resolution & Duration (s)& Subjects & Mode\\
    \midrule
    \textbf{CRCNS}\cite{itti2004crcns}    & 2004 & 50 & $640\times480$ & 6-94 & 4.7(ave) & task \\
    \textbf{SFU}\cite{Hadizadeh2012SFU}   & 2012 & 12 & $352\times288$ & 3-10 & 15 & free\\
    \textbf{DIEM}\cite{Mital2011DIEM}     & 2011 & 84 & $\leq1280\times720$ & 27-217 & 50(ave) & free\\
    \textbf{Hollywood}\cite{Mathe2015pami}   & 2015 & 1857 & $\leq720\times480$ & 2-90 & 19 & task\\
    Xu\cite{xu2017learning}          & 2016 & 32 & $\leq1920\times1080$ & 6-25 & 32 & free\\
    IRCCyN\cite{Boulos2009IRRC1}     & 2009 & 51 & $720\times576$ & 8-10 & 37 & free \\
    VAGBA\cite{Li2011USC}              & 2011 & 50 & $1920\times1080$ & 10 & 14 & free \\
    GazeCom\cite{dorr2010variability}    & 2010 & 18 & $1280\times720$ & 20 & 54 & free \\
    ASCMN\cite{riche2012ascmn}           & 2012 & 24 & $\leq704\times576$ & 2-76 & 13 & free \\
    Coutrot-2\cite{coutrot2015conversation}    & 2015 & 15 & $1232\times504$ & 20-80 & 40 & free \\
    CAMO\cite{Nguyen2013cameramotion} & 2013 & 120 & $640\times480$ & 1.7-8.6 & 10 & free \\
    Marat\cite{marat2007database}     & 2007 & 53 & $720\times576$ & 1-3 & 15 & free \\
    SAVAM\cite{Gitman2014semiauto}    & 2014 & 43 & $1920\times1080$ & 18.1 (ave) & 50 & free \\
    TUD\cite{Alers2012TUD}             & 2012 & 50 & $1280\times720$ & 20 & 12 & task \\
    Peters\cite{peters2007videogame}   & 2007 & 24 & $640\times480$ & 267 & 5 & task \\
    Coutrot-1\cite{Coutrot2013database1} & 2013 & 60 & $720\times576$ & 10-24.8 & 20 & task \\
    \bottomrule
    \end{tabular}%
  \label{tab:addlabel}%
\end{table}%

As discussed in Section \ref{reviewdatabase}, video saliency prediction may benefit from the recent development of deep learning. Unfortunately, as seen in Table \ref{DatabaseDetail}, the existing databases for video saliency prediction are lack of sufficient eye-tracking data to train DNN. Although Hollywood \cite{Mathe2015pami} has 1857 videos, it mainly focuses on task-driven visual saliency. Besides, the video content of Hollywood is limited, only involving human actions of movies. In fact, a large-scale eye-tracking database for video should have 3 criteria: 1) a large number of videos, 2) sufficient subjects, and 3) various video content. In this paper, we establish a large-scale eye-tracking database of videos, satisfying the above three criteria. The detail of our large-scale databases is to be discussed in Section \ref{database}.

\begin{table*}[t]
 \begin{center}
\begin{threeparttable}

  \caption{\footnotesize Categories of videos in our database based on content.}\label{Videocataloge}
    \begin{tabular}{|c|c|c|c|c|c|c|c|}
    \hline
    &\multicolumn{4}{|c|}{Human}&\multirow{2}{*}{Animal}&\multirow{2}{*}{Man-made object}&\multirow{2}{*}{\textbf{Overall}}\\
    \cline{2-5}
    &Daily action&Sports&Social activity&Art performance&&&\\
    \hline
    Number of sub-classes\tnote{*}&23&17&21&19&51&27&148\\
    \hline
    Number of videos&74&58&69&59&156&122&538\\

    \hline
    \end{tabular}%
    \begin{tablenotes}\scriptsize
      \item[*] Here, we also report the number of sub-class in each class of videos. For example, in animal videos, the sub-classes include penguin, rabbit, elephant, etc.
\end{tablenotes}
\end{threeparttable}%
\end{center}
\end{table*}

\begin{figure*}[t]
\begin{center}
\includegraphics[width=.95\linewidth]{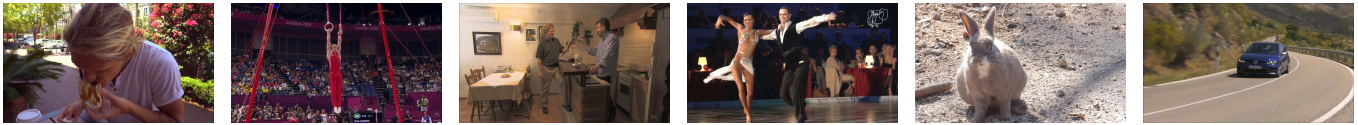}
  \end{center}
\caption{\footnotesize Examples for each class of videos in LEDOV. From left to right, the videos belong to daily action, sports, social activity, art performance, animal and man-made object.}\label{subexample}
\end{figure*}

\section{Dtabase}\label{database}

In this section, a new Large-scale Eye-tracking Database of Videos (LEDOV) is established, which is available online for facilitating the future research. More details and analysis about our LEDOV database are discussed in the following.

\begin{table}[h]\scriptsize
   \begin{center}
  \caption{\footnotesize Numbers of videos with different amount of objects.}\label{Videocataloge2}
       \begin{tabular}{|c|c|c|c|c|}
    \hline
     Object number&$=1$&$=2$&$>2$&\textbf{Overall}\\
    \hline
    Video amount&377&103&58&538\\
    \hline
    \end{tabular}%
    \end{center}
\end{table}%
\subsection{Databases establishment}\label{database_establish}
We present our LEDOV database from the aspects of stimuli, apparatus, participant and procedure.

\textbf{Stimuli.}
538 videos, in total 179,336 frames and 6,431 seconds, were collected, according to the following four criteria.
 \begin{enumerate}[]
    \item Diverse video content. Videos with diverse content were sourced from different online accessible repositories, such as daily vlogs, documentaries, movies, sport casts, TV shows, etc. In Table \ref{Videocataloge}, we briefly classify all 538 videos according to their content. Some examples are shown in Figure \ref{subexample}.
    \item Including at least one object. Only videos with at least one object were qualified for our database. Table \ref{Videocataloge2} reports the numbers of videos with different amount of objects in our database.
    \item High quality. We ensured high quality of videos in our database by choosing those with at least 720p resolution and 24 Hz frame rate. To avoid quality degradation, the bit rates were maintained when converting videos to the uniform MP4 format.
    \item Stable shot. The videos with unsteady camera motions and frequent cinematic cuts were not included in LEDOV. Specifically, there are 212 videos with stable camera motion. Other 316 videos are without any camera motion.
 \end{enumerate}

\textbf{Apparatus\&Participants.}
For monitoring the binocular eye movements, an eye tracker, Tobii TX300 \cite{Tobiitx300}, was used in our experiment. TX300 is an integrated eye tracker with a 23'' TFT monitor at screen resolution of $1920\times1080$. During the experiment, TX300 captured gaze data at 300 Hz. According to \cite{Tobiitx300}, the gaze accuracy can reach 0.4 vision degree (around 15 pixels in stimuli) in the ideal working condition\footnote{The ideal condition is that the illumination in working environment is constant at 300 lux, and that the distance between subjects and the monitor is fixed at 65 cm. Such condition was satisfied in our eye-tracking experiment.}. For more details about Tobii TX300, refer to \cite{Tobiitx300}. Moreover, 32 participants (18 males and 14 females), aging from 20 to 56 (32 in average), were recruited to take part in the eye-tracking experiment. All participants were non-expert for the eye-tracking experiment, with normal/corrected-to-normal vision. It is worth pointing out that only those who passed the calibration of the eye tracker and had less than $10\%$  fixation dropping rate, were quantified for our eye tracking experiment. As a result, 32 among 60 subjects were selected in our experiment.


\textbf{Procedure.}
Since visual fatigue may arise after viewing videos for a long time, 538 videos in LEDOV were equally divided into 6 non-overlapping groups with similar numbers of videos in content (i.e., human, animal and man-made object). During the experiment, each subject was seated on an adjustable chair around 65 cm from the screen, followed by a 9-point calibration. Then, the subject was required to free view 6 groups of videos in a random order. In each group, the videos were also displayed at random. Between two successive videos, we inserted a 3-second rest period with black screen and a 2-second guidance image with a red circle in screen center. As such, the eyes can be relaxed, and then the initial gaze location can be reset at center. After viewing a group of videos, the subject was asked to take a break until he/she was ready for viewing the next group of videos. Finally, 5,058,178 fixations (saccades and other eye movements have been removed) were recorded from 32 subjects on 538 videos, for our LEDOV database.

%

\subsection{Database analysis}\label{database_analysis}
In this section, we mine our database to analyze human attention on videos. More details are introduced as follows.

\subsubsection{Temporal correlation of attention on consecutive frames}\label{ccframe}
\begin{figure}
    \centering
      \subfigure[\tiny 2s before]{\includegraphics[width=.19\linewidth]{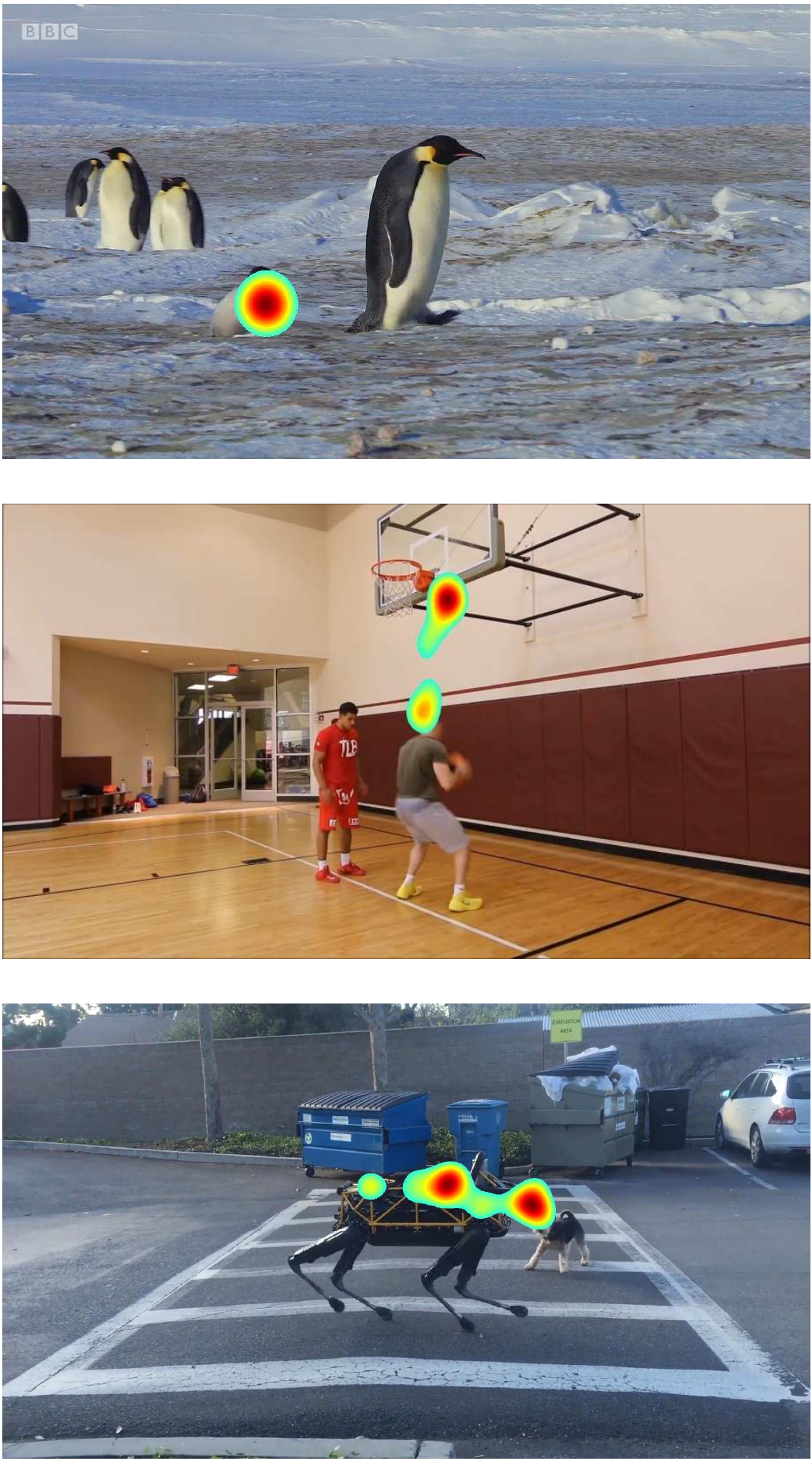}}
      \subfigure[\tiny 1.5s before]{\includegraphics[width=.19\linewidth]{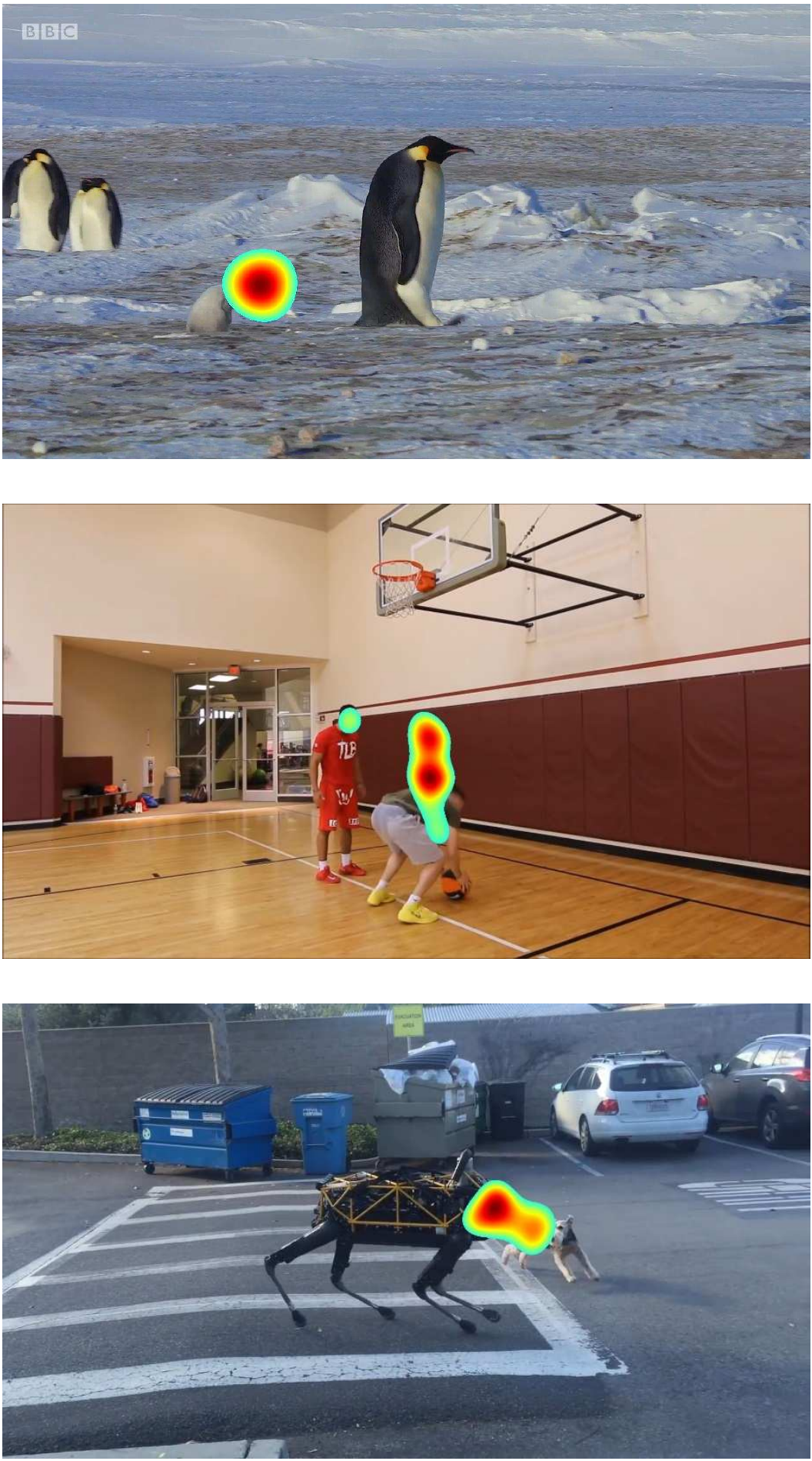}}
      \subfigure[\tiny 1s  before]{\includegraphics[width=.19\linewidth]{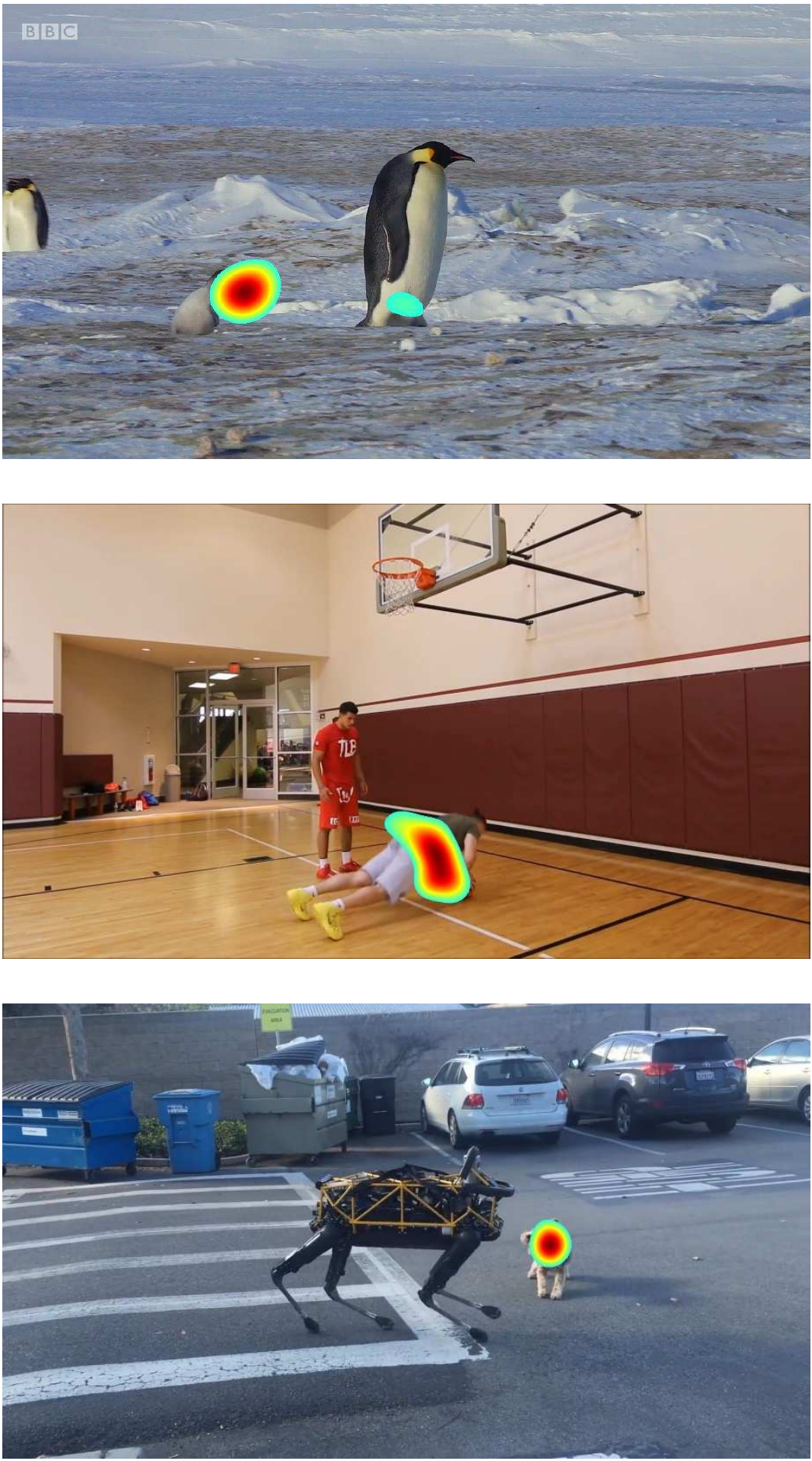}}
      \subfigure[\tiny 0.5s before]{\includegraphics[width=.19\linewidth]{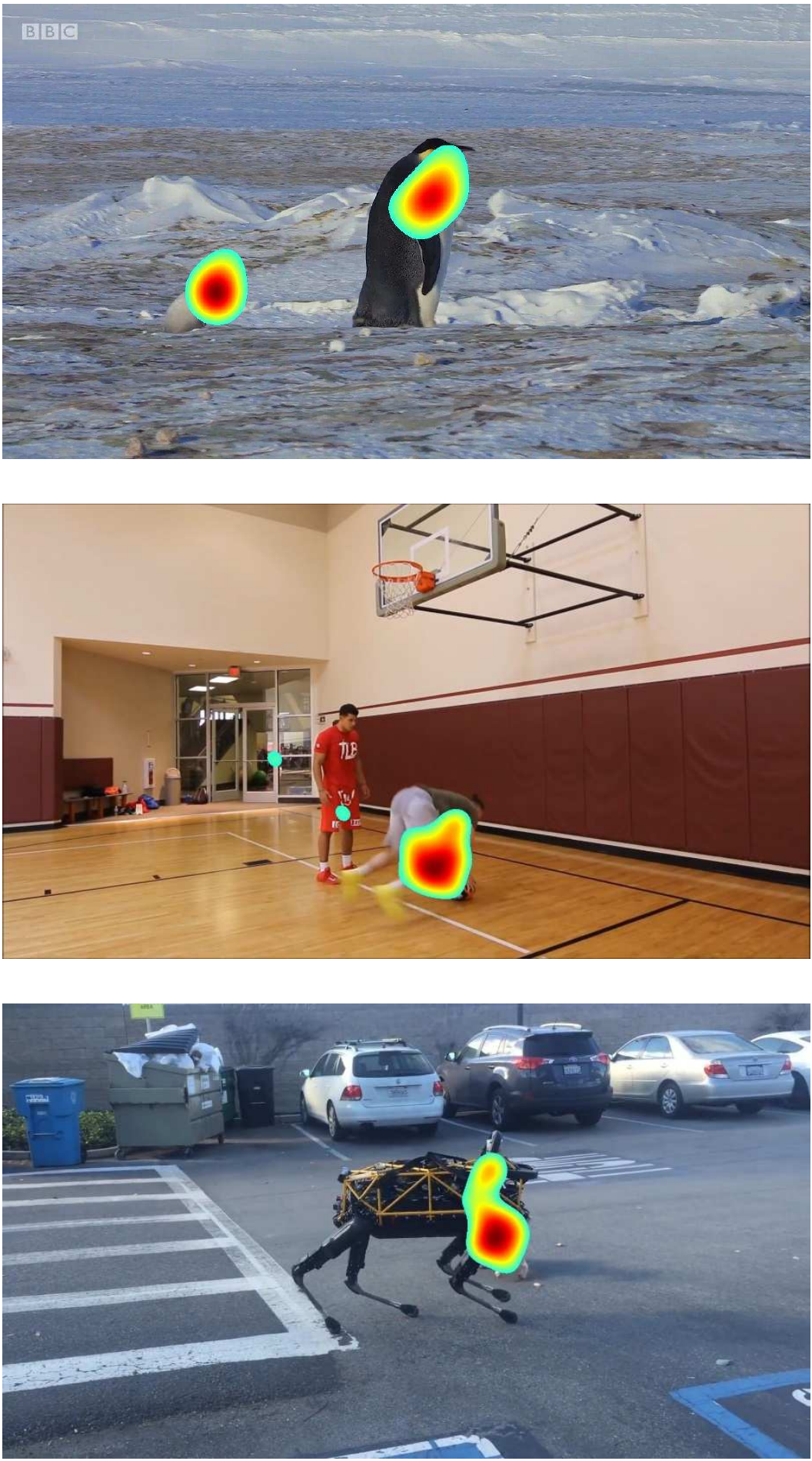}}
      \subfigure[\tiny Current frame]{\includegraphics[width=.19\linewidth]{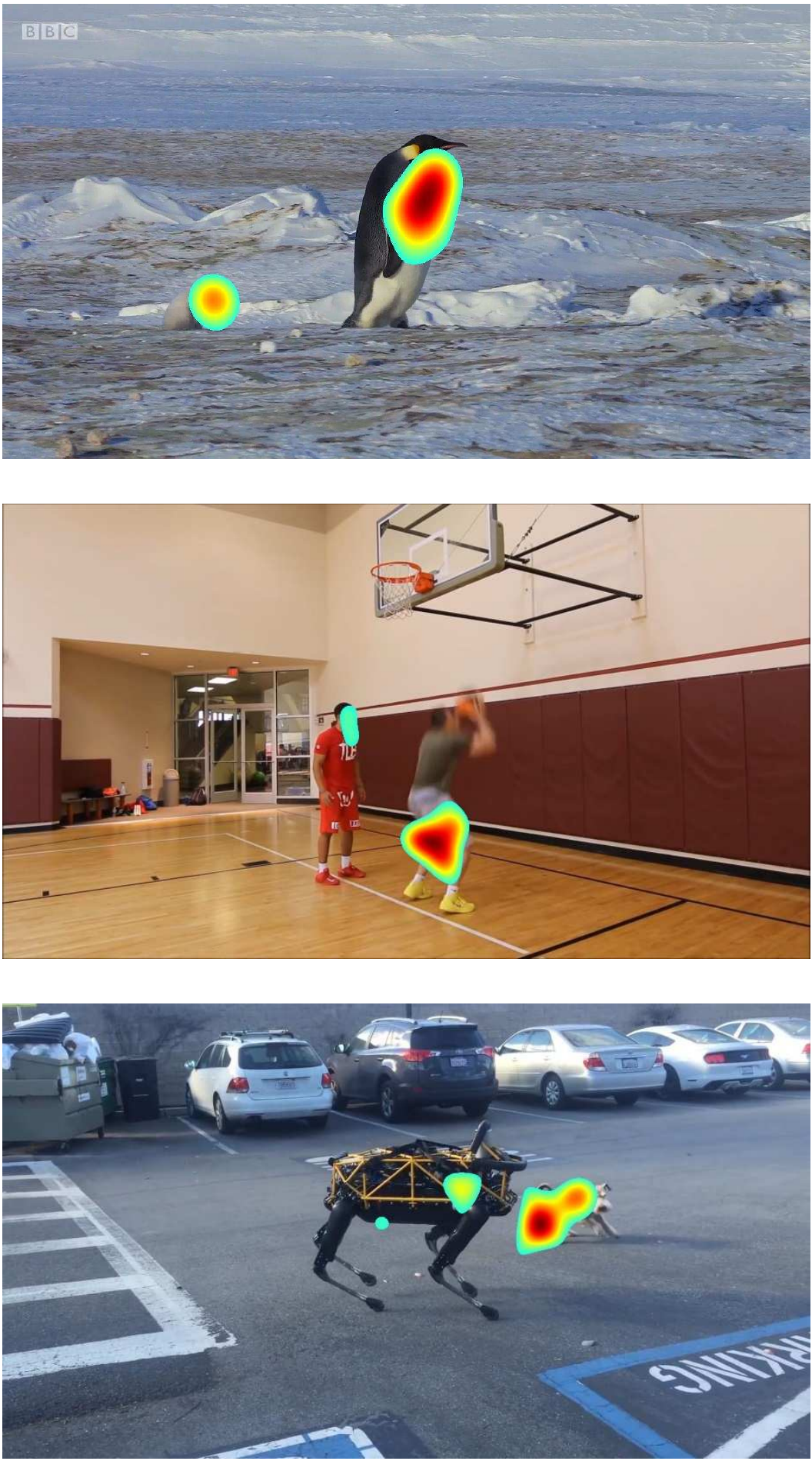}}
    \caption{\footnotesize Examples for human fixation maps across consecutive frames.}
    \label{consecutiveframes}
\end{figure}

\begin{figure}
\begin{center}
\includegraphics[width=.95\linewidth]{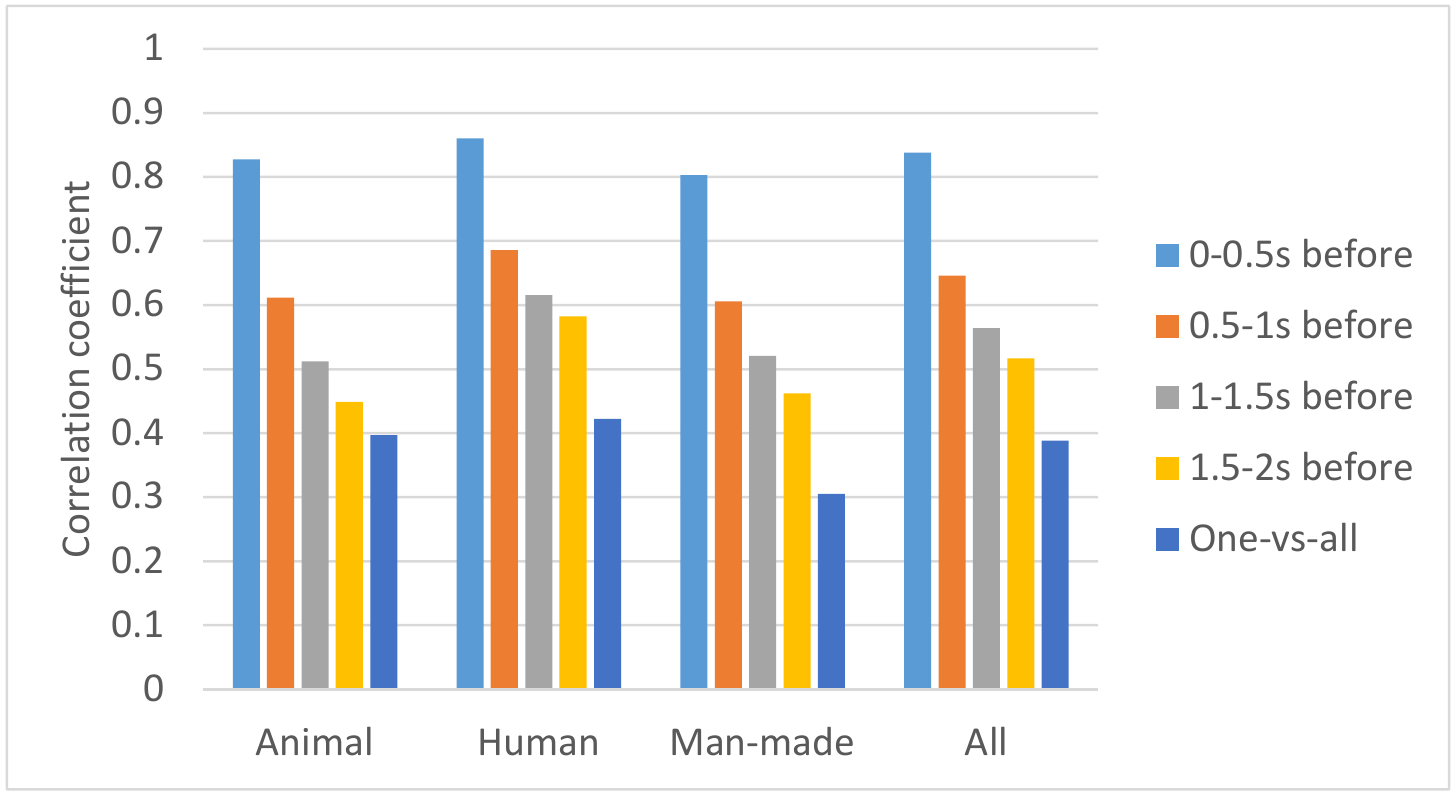}
  \end{center}
\caption{\footnotesize The CC results of temporal attention correlation, averaged over animal, human, man-made object and all videos in LEDOV. }\label{cc_result}
\end{figure}

It is interesting to explore the temporal correlation of attention across consecutive frames. In Figure \ref{consecutiveframes}, we show human fixation maps along with some consecutive frames, for 3 selected videos. As we can see from Figure \ref{consecutiveframes}, there exists high temporal correlation of attention across consecutive frames of videos.  To quantify such correlation, we further measure the linear correlation coefficient (CC) of fixation map between two consecutive frames. Assume that $\mathbf{G}_c$ and $\mathbf{G}_p$ are fixation maps of current and previous frames. Then, the CC value of fixation maps averaged over a video can be calculated as follows,


\begin{eqnarray}
&&CC=\frac{1}{|\mathbf{V}_c|}\sum_{c\in \mathbf{V}_c}\frac{1}{|\mathbf{V}_p|}\sum_{p\in \mathbf{V}_p}\frac{\rm{Cov}(\mathcal{N}(\mathbf{G}_c),\mathcal{N}(\mathbf{G}_p))}{\rm{Std}(\mathcal{N}(\mathbf{G}_c))\cdot\rm{Std}(\mathcal{N}(\mathbf{G}_p))},\nonumber\\
&&\rm{where} \qquad \mathcal{N}(\mathbf{G}_c)= \frac{\mathbf{G}_c-\rm{Mean}(\mathbf{G}_c)}{\rm{Std}(\mathbf{G}_c)}.\label{ccave2}
\end{eqnarray}
In \eqref{ccave2}, $\mathbf{V}_c$ is the set of all frames in the video, while $\mathbf{V}_p$ is the set of consecutive frames before frame c. Additionally, $\rm{Cov(\cdot)}$, $\rm{Std(\cdot)}$ and $\rm{Mean(\cdot)}$ are covariance, standard deviation and mean operators. For $\mathbf{V}_p$, we choose 4 sets of previous frames, i.e., 0-0.5s before, 0.5-1s before, 1-1.5s before and 1.5-2s before. Then, in Figure \ref{cc_result}, we plot the CC results of these 4 sets of $\mathbf{V}_p$, which are averaged over all videos in our LEDOV database. We also show in Figure \ref{cc_result} one-vs-all results, which is the baseline of averaged CC between fixation maps of one subject and the rest (indicating attention correlation between humans). We can see from this figure that the  CC value of temporal consistency is much higher than that of one-vs-all baseline. This implies high temporal correlation across consecutive frames of video. We can further find that temporal correlation of attention decreases, when increasing the distance between the current and previous frames. Consequently, the long- and short-term dependency of attention across frames of video can be verified.


\subsubsection{Correlation between objectness and human attention}\label{objectness}
\begin{figure}
    \centering
      \subfigure[\tiny Current frame]{\includegraphics[width=.23\linewidth]{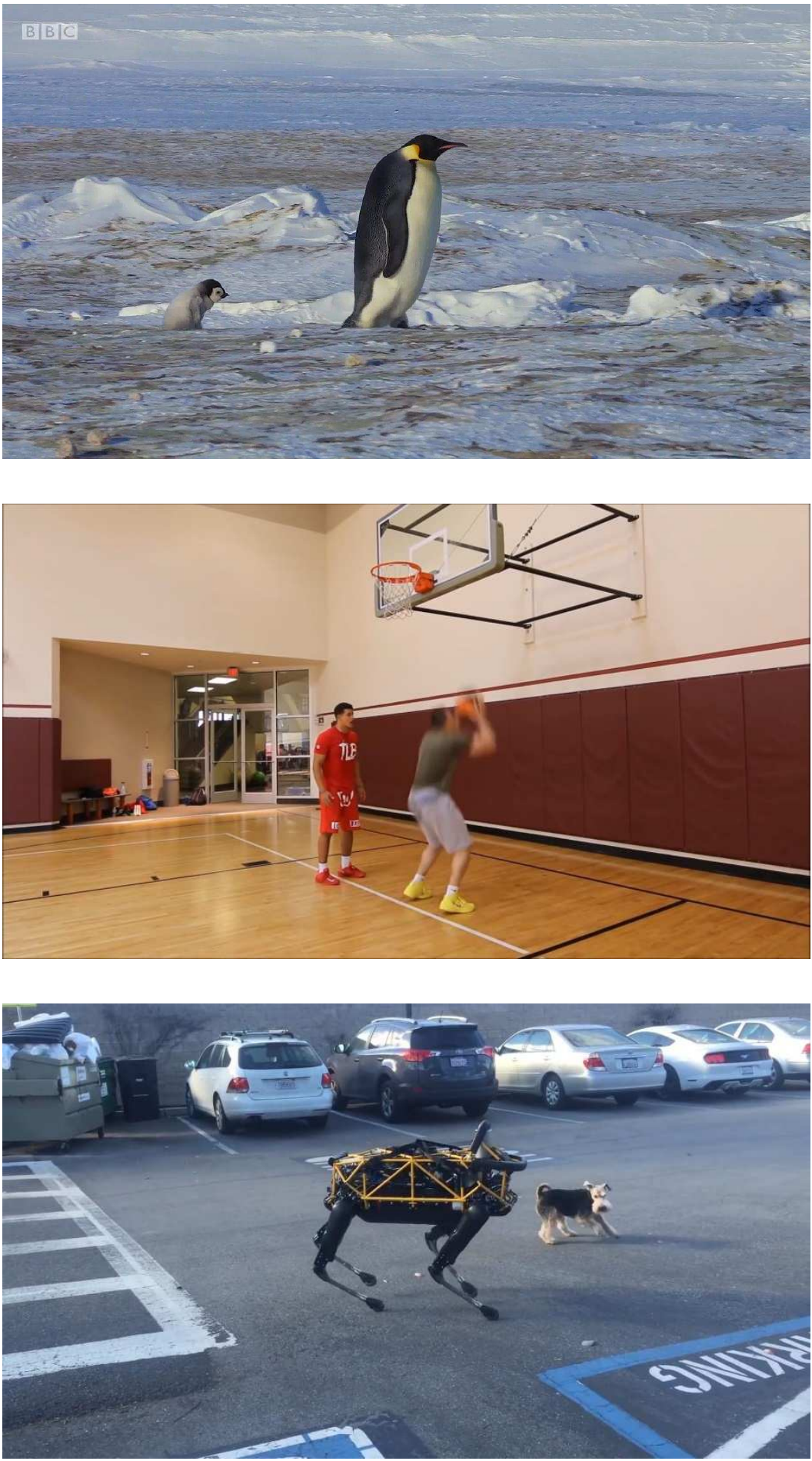}}
      \subfigure[\tiny One candidate]{\includegraphics[width=.23\linewidth]{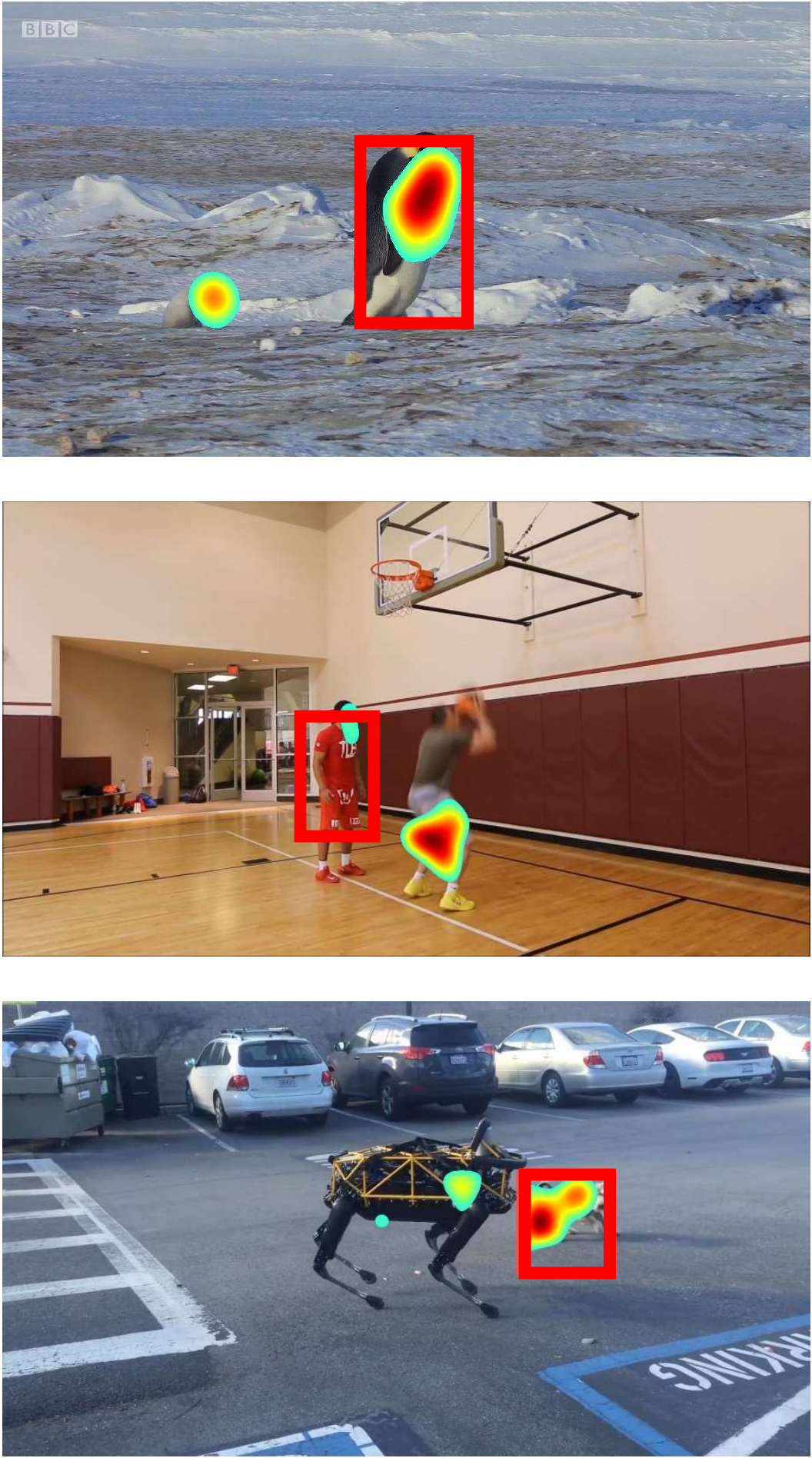}}
      \subfigure[\tiny Multiple candidates]{\includegraphics[width=.23\linewidth]{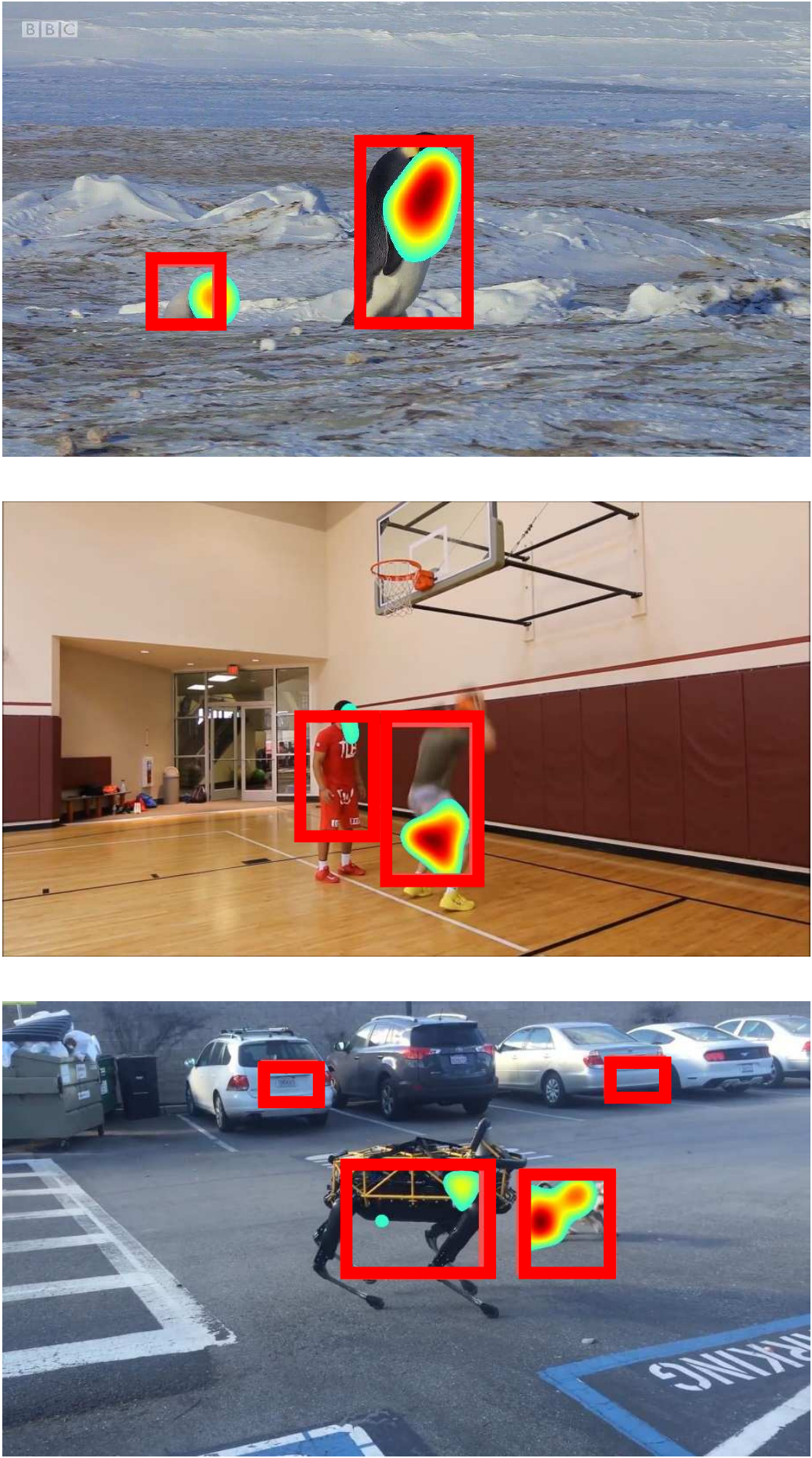}}
      \subfigure[\tiny Optical flow]{\includegraphics[width=.23\linewidth]{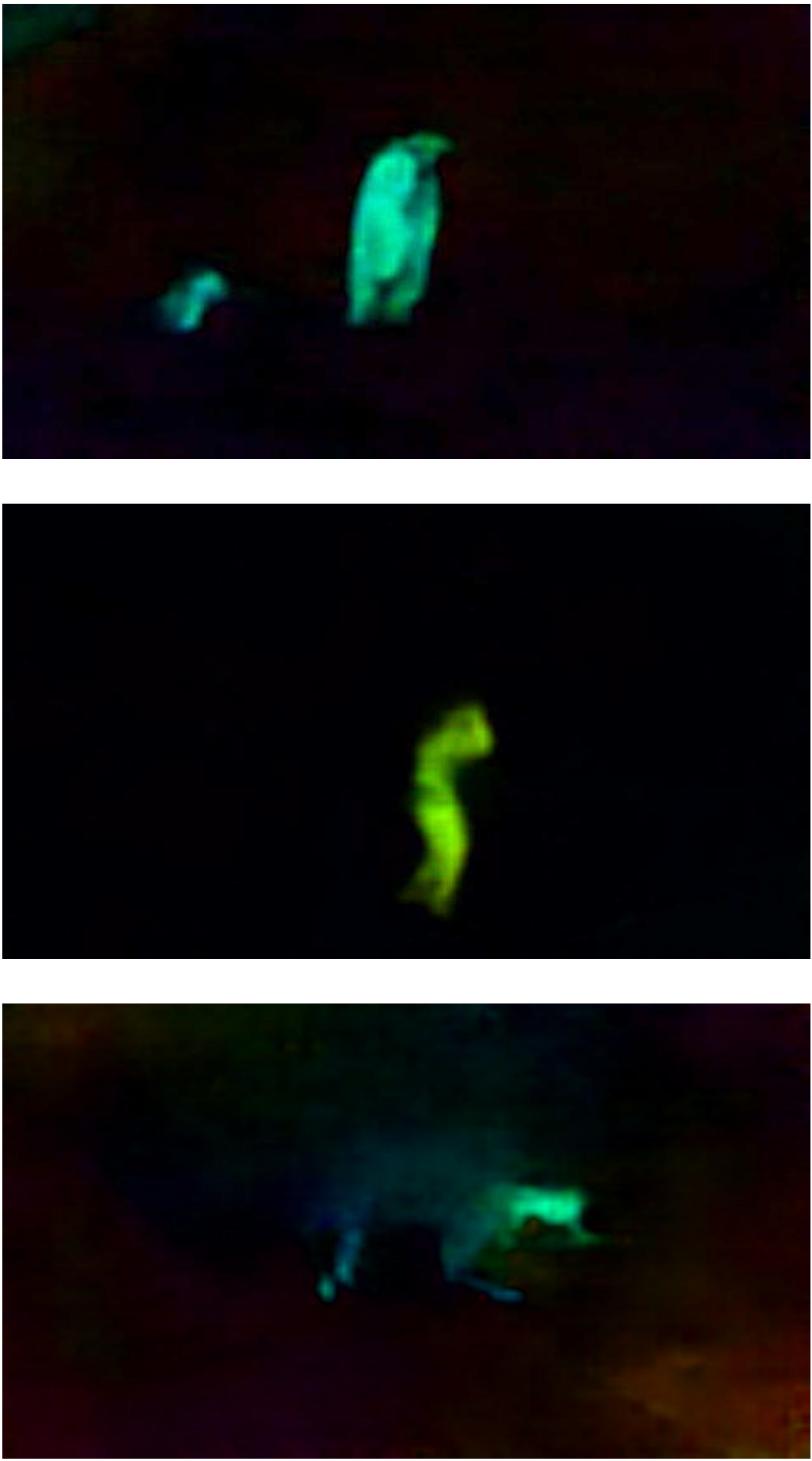}}
        \caption{\footnotesize Examples of ground-truth fixation maps and candidate objects detected by YOLO. (a) shows randomly selected frames from three videos in our LEDOV database. (b) illustrates the fixation maps as well as one candidate object in each frame. (c) demonstrates fixation maps and multiple candidate objects. (d) displays optical flow maps of each frame, represented in HSV color space.}
    \label{dataExample}
\end{figure}

\begin{figure*}
\begin{center}
\subfigure[Fixation hit proportion]{\includegraphics[width=.4\linewidth]{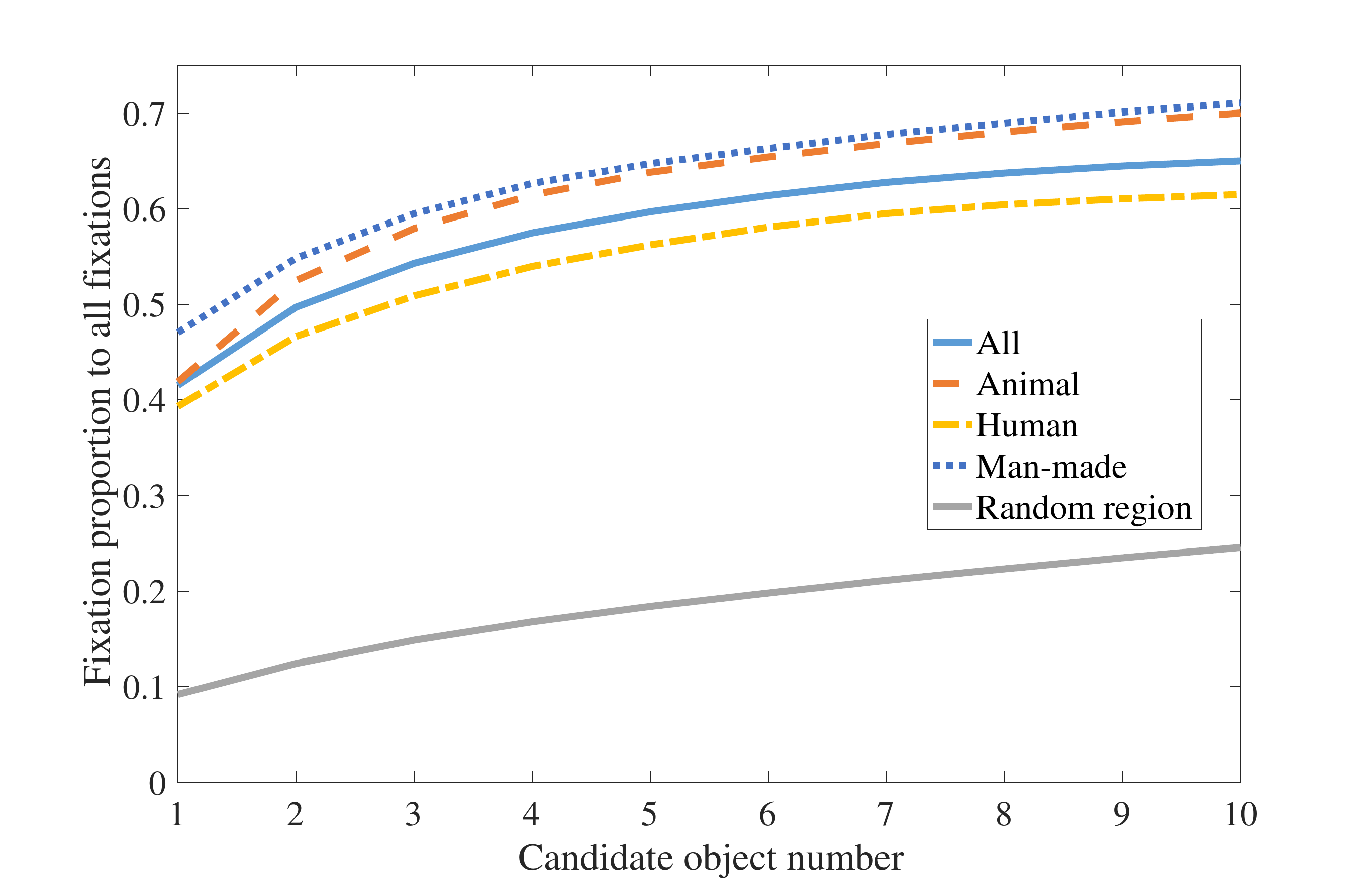}}
\subfigure[Fixation area proportion]{\includegraphics[width=.4\linewidth]{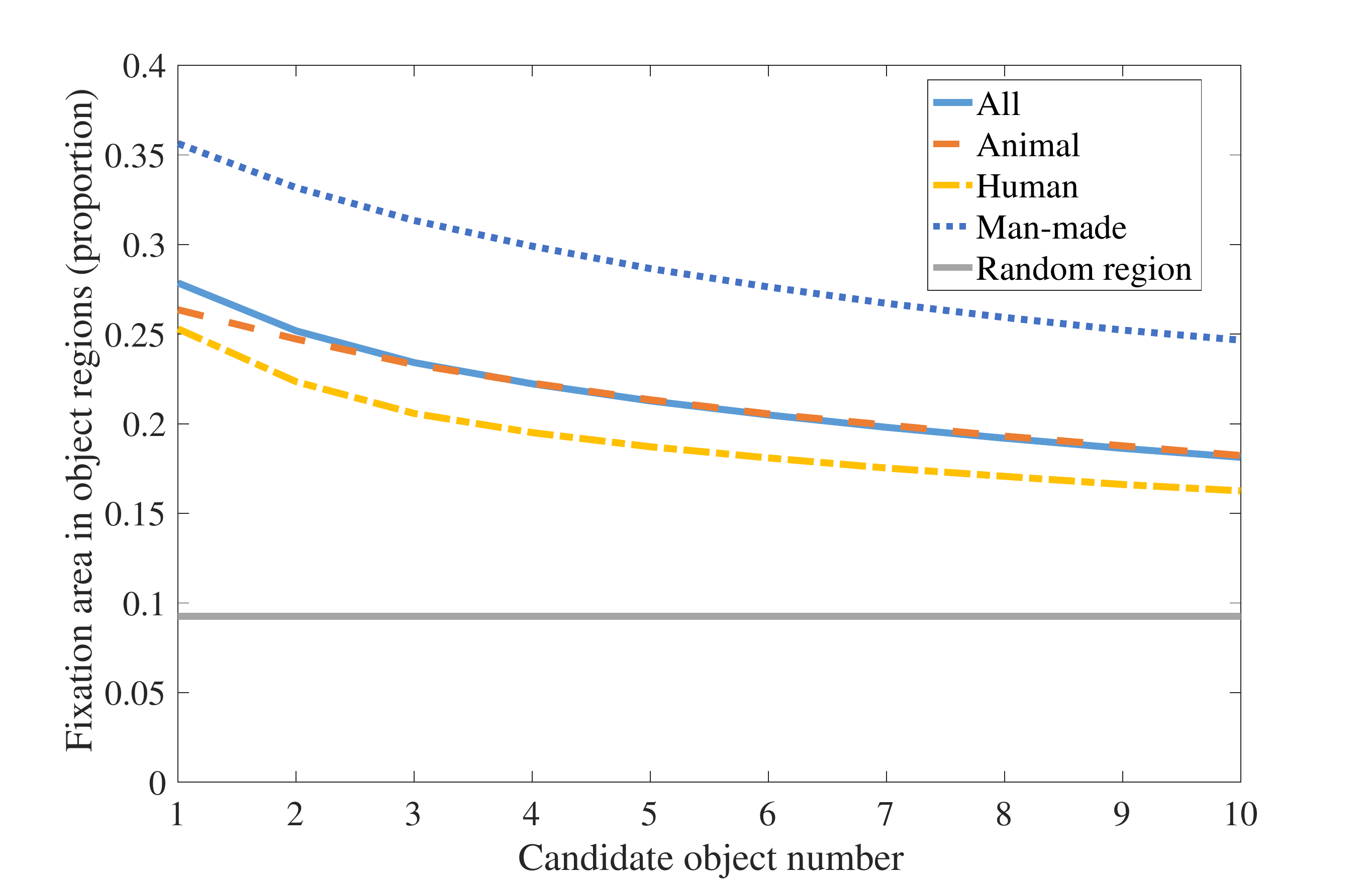}}
  \end{center}
\caption{\footnotesize (a) Fixation proportion belonging to object regions at increased numbers of detected candidate objects. (b) Proportion of fixation area in object regions along with increased numbers of detected candidate objects. The results of all videos as well as animals, human and man-made object videos are plotted with different curves. Besides, the results of fixations hitting on random region are plotted as the baseline.}\label{spacialanalysis}
\end{figure*}

It is intuitive that people may be attracted by objects rather than background when watching videos. Therefore, we investigate how much attention is related to object regions. First, we apply a CNN-based objection detection method YOLO \cite{Redmon2015} to detect main objects in each video frame. Here, we generate different numbers of candidate objects in YOLO, via setting thresholds of confidence probability and non-maximum suppression. Figure \ref{dataExample}-(b) shows the examples of one detected object, while Figure \ref{dataExample}-(c) shows the results for more than one object. We can observe from Figure \ref{dataExample}-(b) that attention is normally attended to object regions. We can also see from Figure \ref{dataExample}-(c) that more human fixations can be included along with increased number of detected candidate objects. To quantify the correlation between human attention and objectness, we measure the proportion of fixations falling into object regions to those of all regions. In Figure \ref{spacialanalysis}-(a), we show the fixation proportion at increased number of candidate objects, averaged over all videos in LEDOV. We can observe from this figure that fixation proportion hitting on object regions is much higher than that hitting on random region. This implies that there exists high correlation between objectness and human attention when viewing videos.  Figure \ref{spacialanalysis}-(a) also shows that fixation proportion increases alongside more candidate objects, which indicates that human attention may be attracted by more than one object.

In addition, one may find from Figure \ref{dataExample} that human attention is attended to only small parts of object regions. Therefore, we measure the proportion of fixation area\footnote{The fixation area is computed by a pre-set threshold on the fixation map. The fixation map is generated by the fixation points convolved with a Gaussian filter as \cite{rajashekar2008gaffe}.} inside the object to the whole object area. Figure \ref{spacialanalysis}-(b) shows the results of such proportion, at increased number of detected candidate objects. We can see from this figure that the proportion of fixation area proportion decreases as at more candidate objects.

%


\subsubsection{Correlation between motion and human attention}\label{opticalflow}

\begin{figure*}
    \centering
      \subfigure{\includegraphics[width=.12\linewidth]{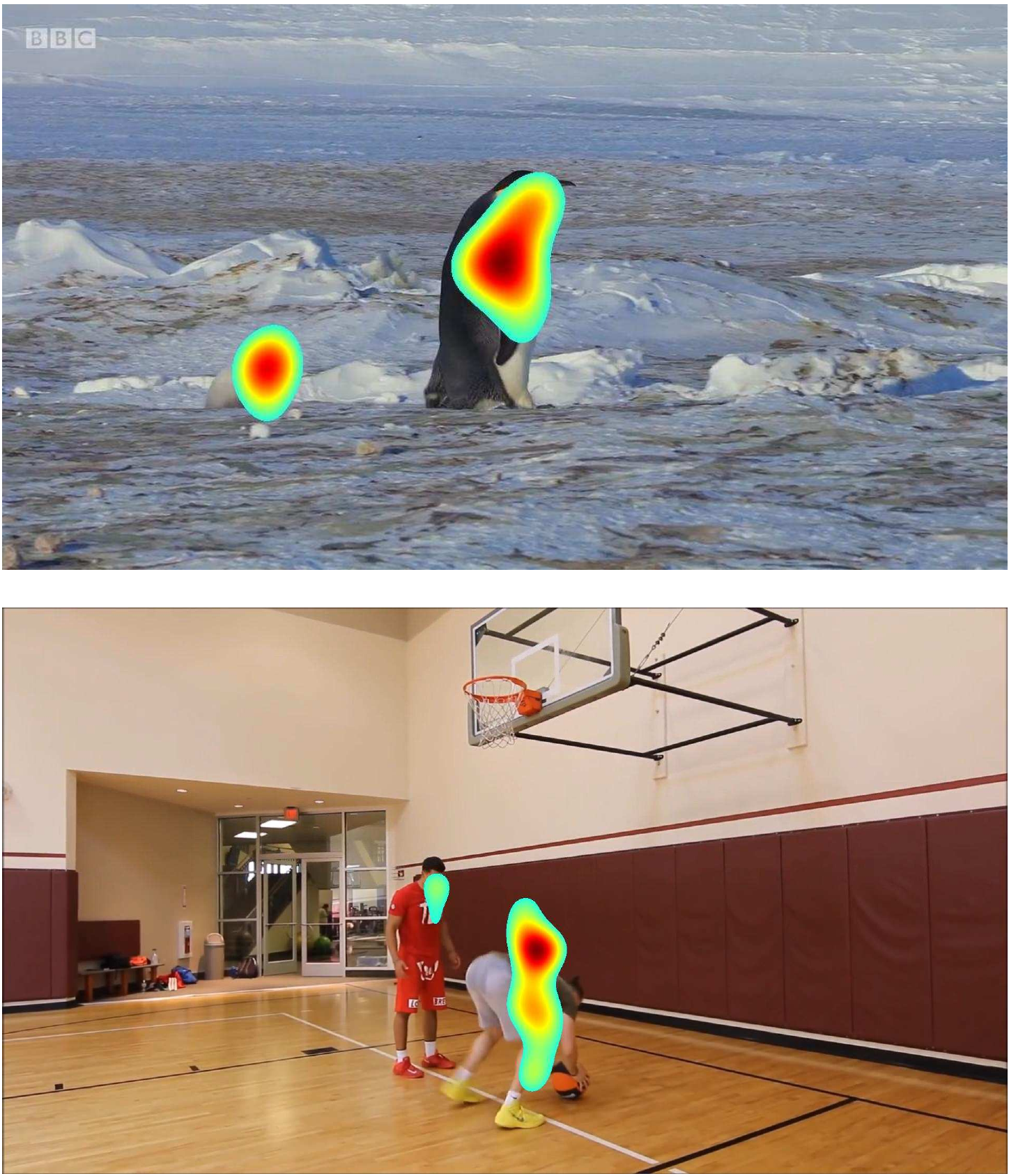}}
      \subfigure{\includegraphics[width=.12\linewidth]{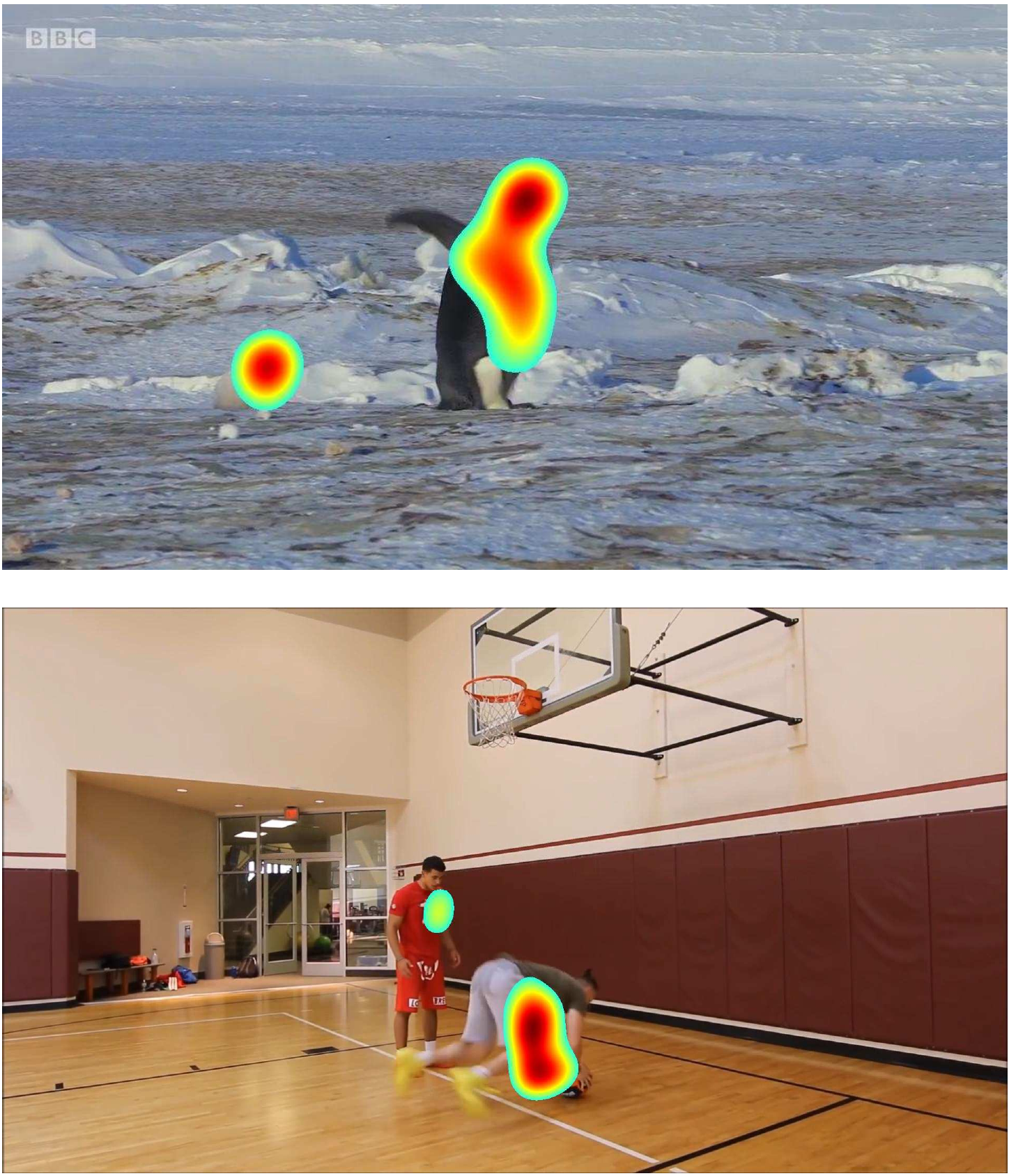}}
      \subfigure{\includegraphics[width=.12\linewidth]{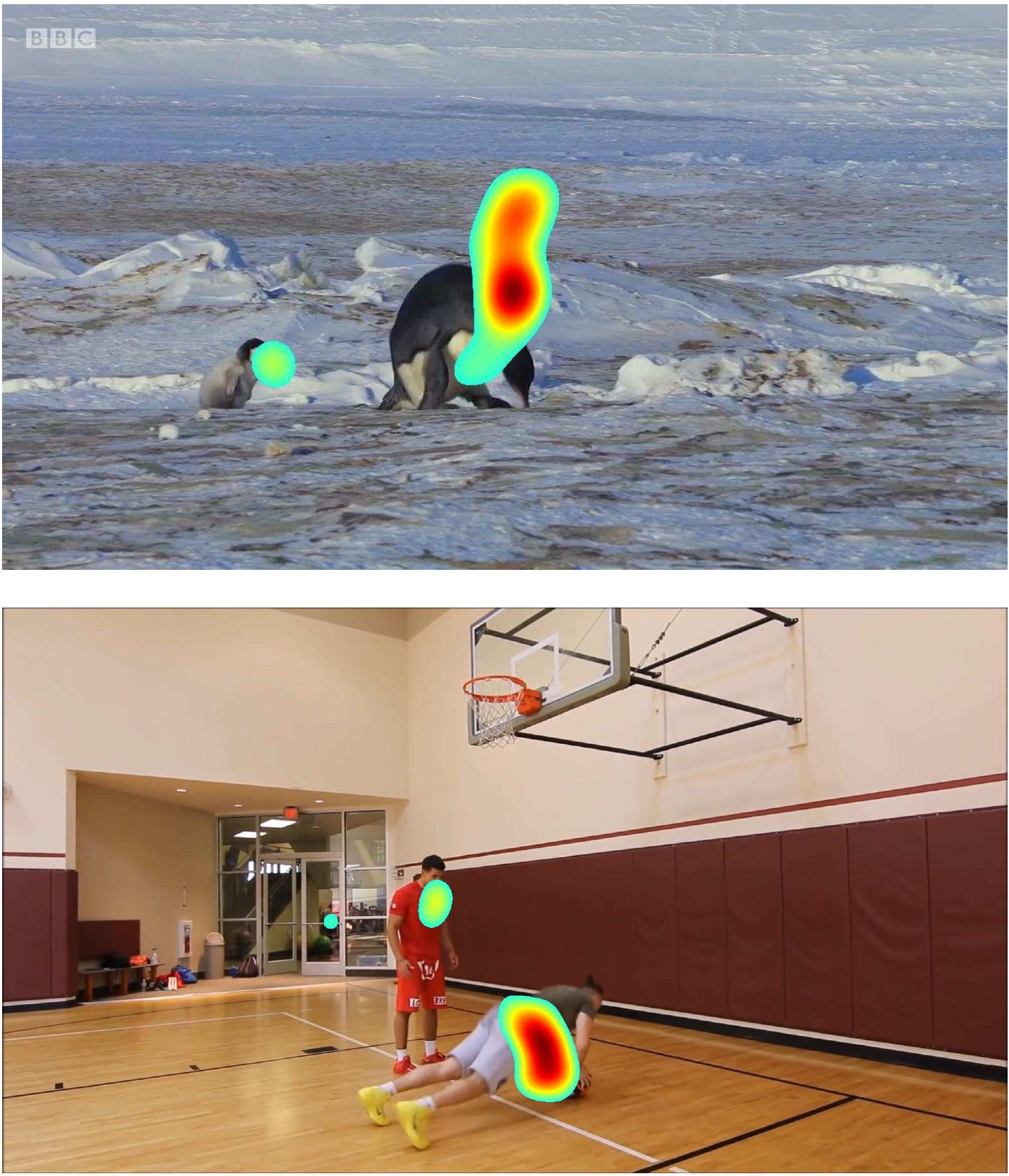}}
      \subfigure{\includegraphics[width=.12\linewidth]{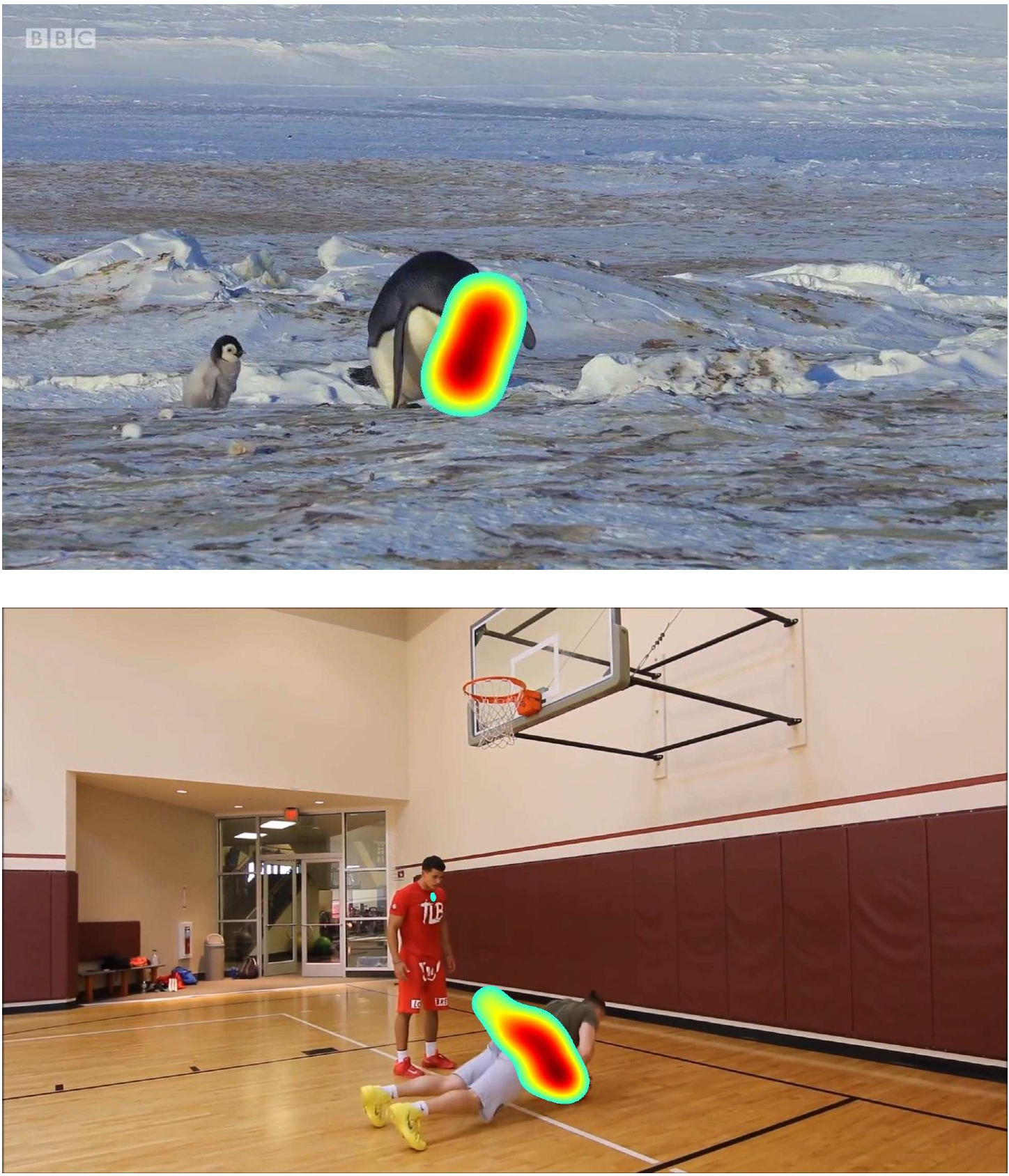}}
      \subfigure{\includegraphics[width=.12\linewidth]{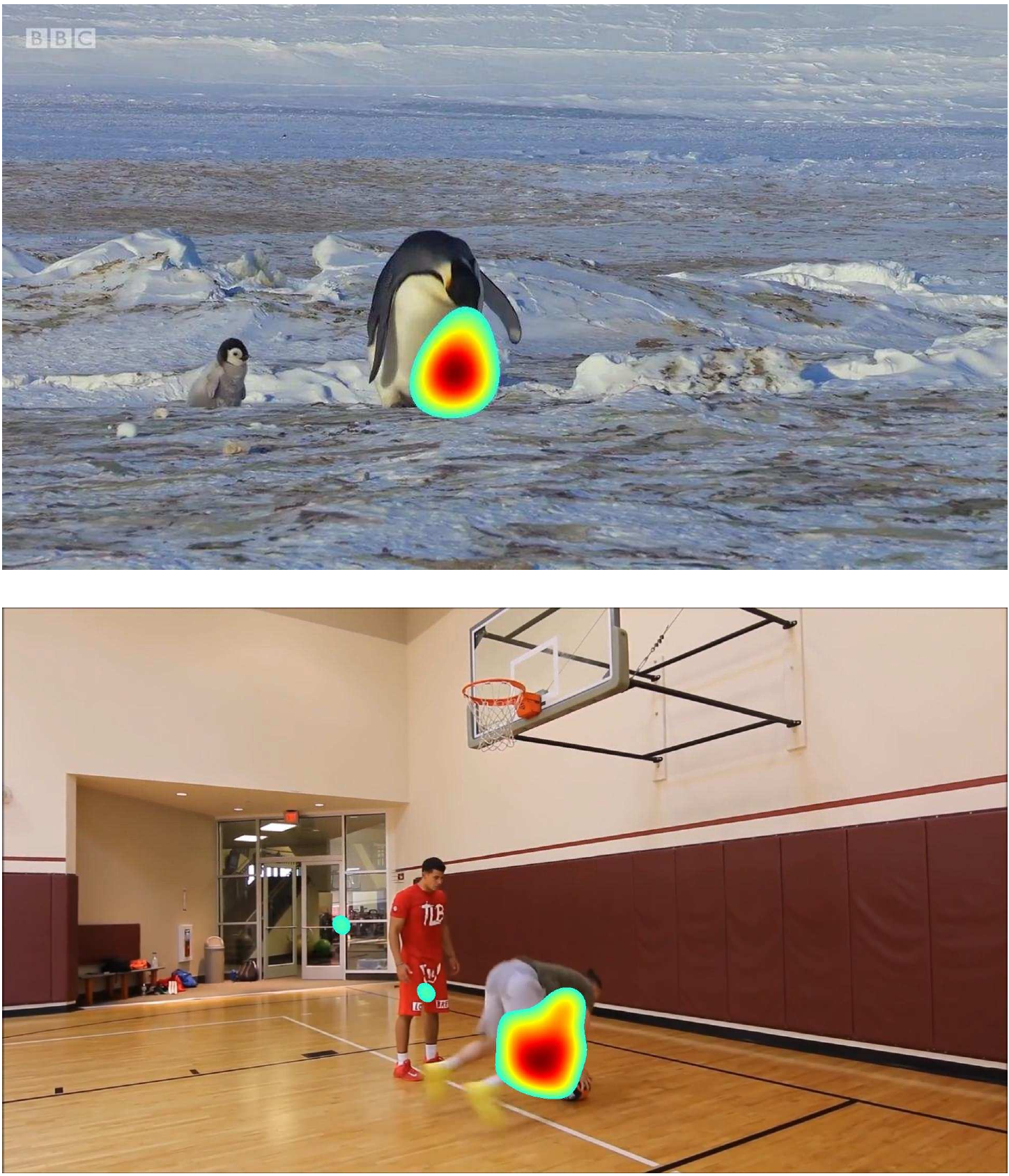}}
      \subfigure{\includegraphics[width=.12\linewidth]{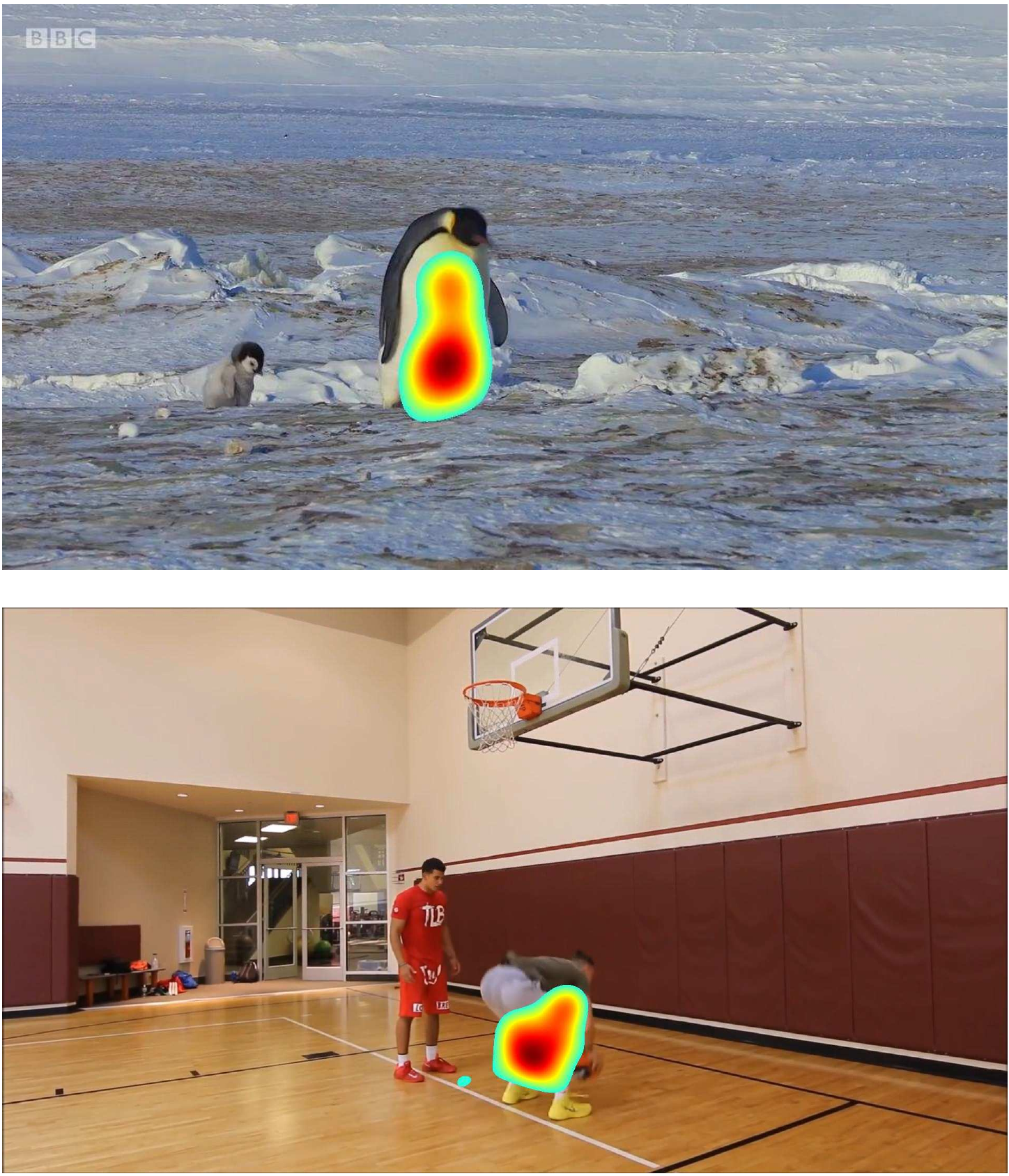}}
    \caption{\footnotesize Attention heat maps of some frames selected from video \emph{animal\_penguin07} and \emph{human\_basketball02}, where the human attention is attracted by moving objects or the moving parts of objects.}
    \label{mvcompare}
\end{figure*}

\begin{figure}
\begin{center}
\includegraphics[width=.7\linewidth]{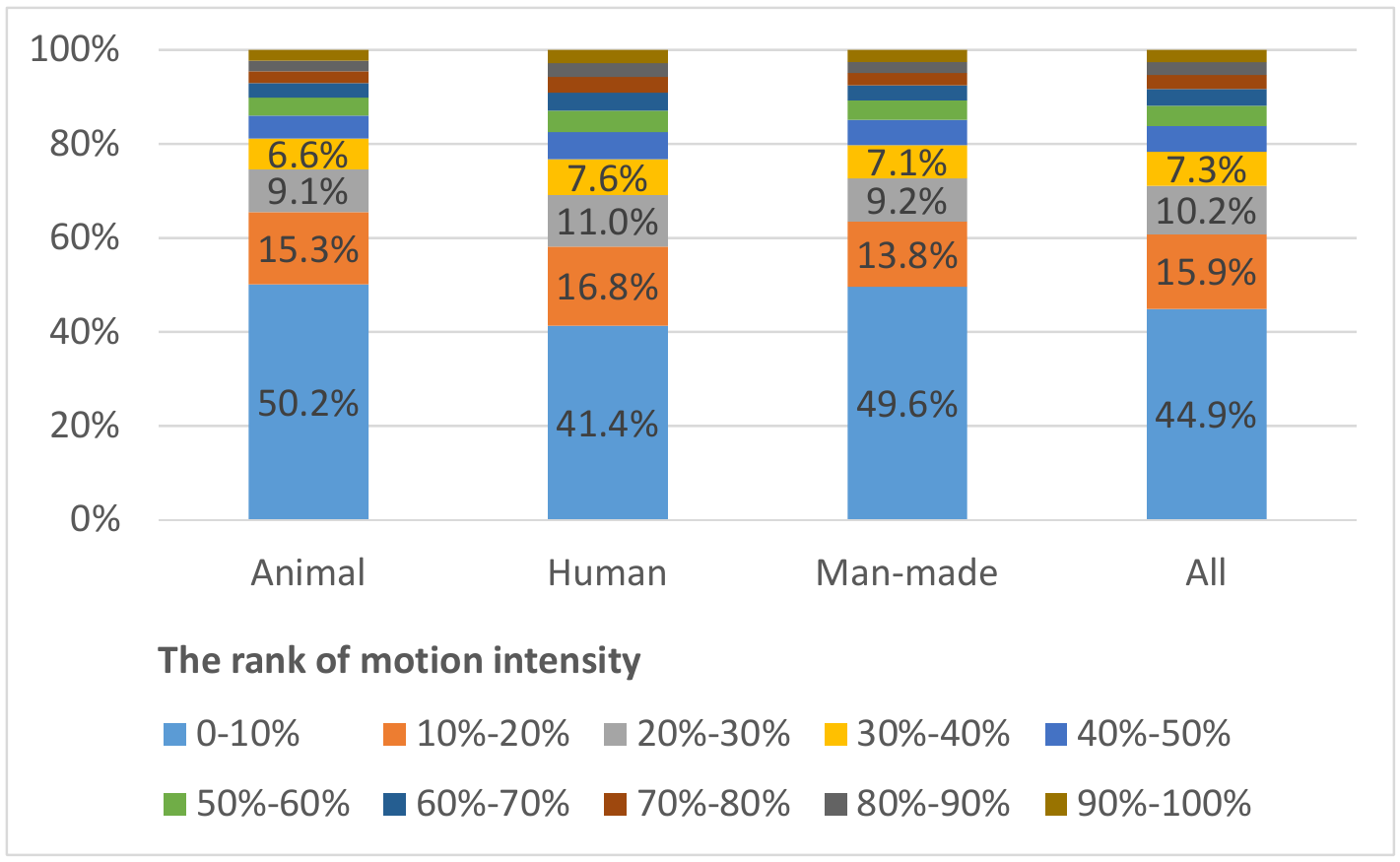}
  \end{center}
\caption{\footnotesize Proportion of fixations belonging to 10 groups, ranked according to the motion intensity. }\label{mvgroup}
\end{figure}

From our LEDOV database, we find that human attention trends to focus on moving objects or the moving parts of objects. Specifically, as shown in the first row of Figure \ref{mvcompare}, human attention is transited to the big penguin, when it suddenly falls with a rapid motion. Besides, the second row of Figure \ref{mvcompare} shows that, in the scene with a single salient object, the intensive moving parts of the player may considerably attract more fixations than other parts. It is interesting to further explore the correlation between motion and human attention inside the regions of objects. Here, we apply FlowNet \cite{dosovitskiy2015flownet}, a CNN based optical flow method, to measure the motion intensity in all frames (some results are shown in Figure \ref{dataExample}-(d)). At each frame, pixels are ranked according to the descending order of motion intensity. Subsequently, we cluster the ranked pixels into 10 groups with equal number of pixels, over all video frames in the LEDOV database. For example, the first group includes pixels with top $10\%$ ranked motion intensity. The numbers of fixations falling into each group are shown in Figure \ref{mvgroup}. We can see from Figure \ref{mvgroup} that $44.9\%$ fixations belong to the group with top $10\%$ high-valued motion intensity. This implies the high correlation between motion and human attention within the region of objects.

%

\section{Proposed method}\label{method}
\subsection{Framework}\label{framework}
\begin{figure*}
    \centering
      \subfigure[The overall architecture of OM-CNN]{\includegraphics[width=.6\linewidth]{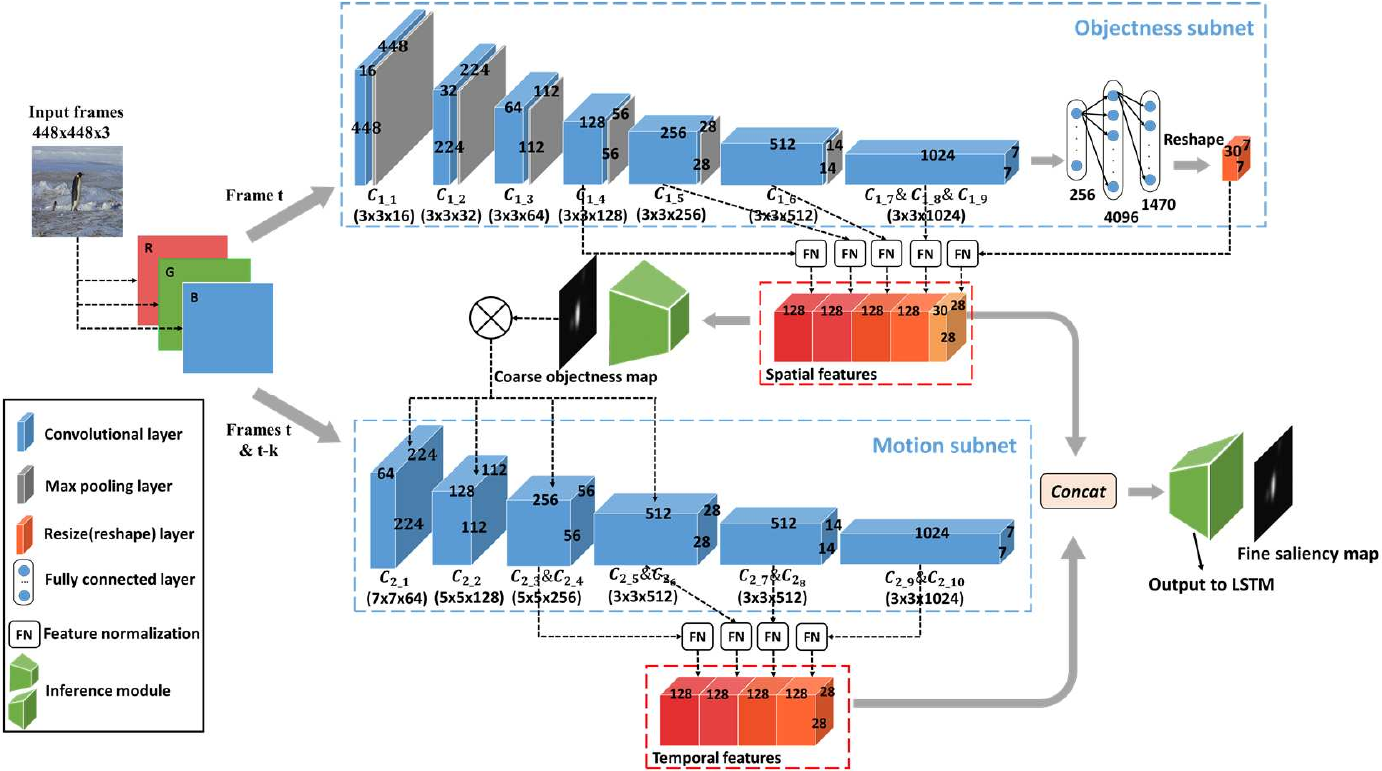}}
      \subfigure[The details for sub-modules of inference module and feature normalization]{\includegraphics[width=.5\linewidth]{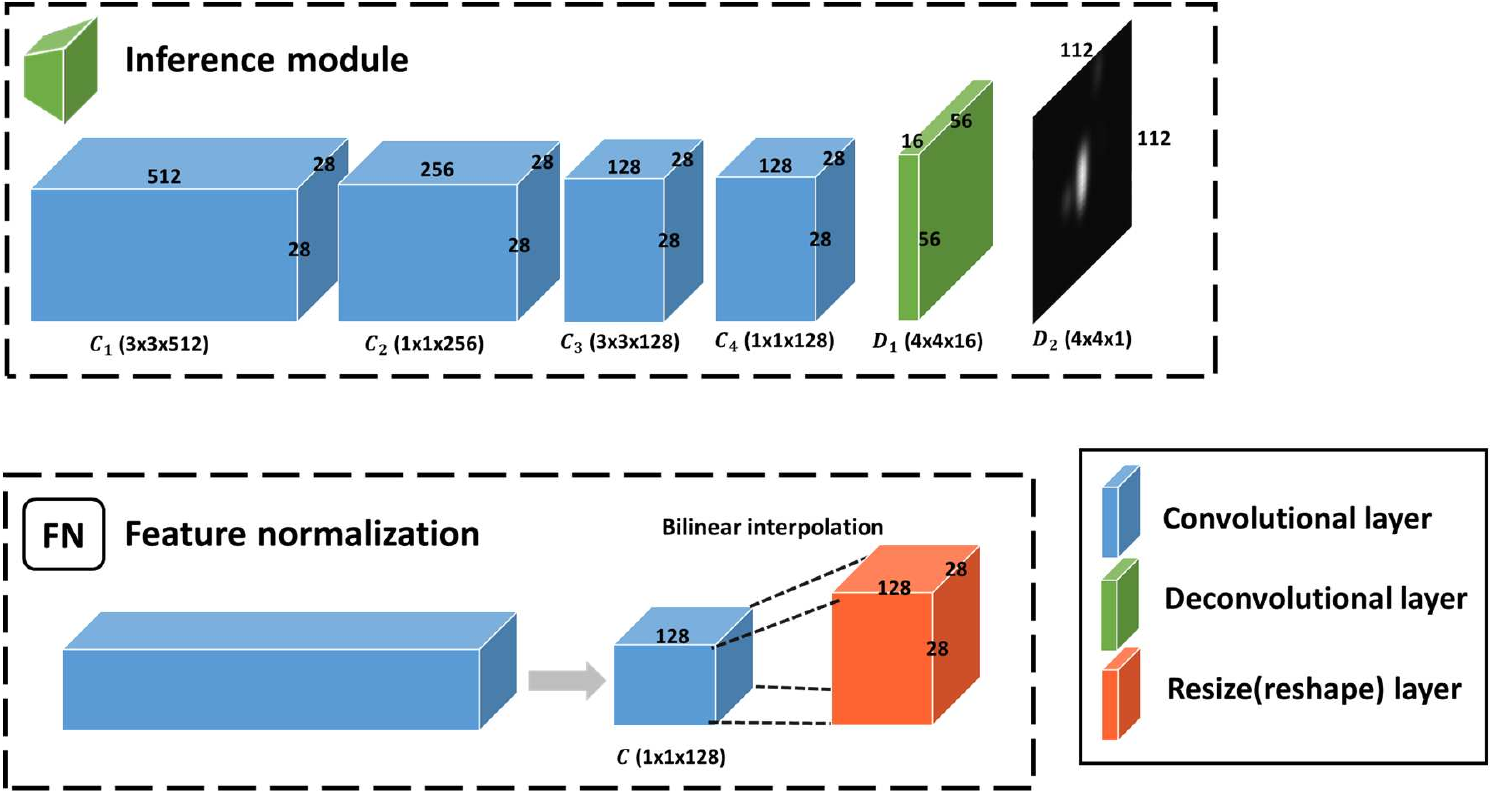}}
    \caption{\footnotesize Overall architecture of our OM-CNN for predicting video saliency of intra-frame. The sizes of convolutional kernels are shown in the figure. For instance, $3\times3\times16$ means 16 convolutional kernels with size of $3\times3$. Note that the $7-9$th \emph{convolutional layers} ($C_{1\_7}$,$C_{1\_8}\&C_{1\_9}$) in the objectness subnet have the same size of convolutional kernels, thus sharing the same cube in (a) but not sharing the parameters. Similarly, each of the last four cubes in the motion subnet represents 2 \emph{convolutional layers} with same kernel size. The details of the inference and feature normalization modules are shown in (b).}
    \label{fOMCNN}
\end{figure*}

For video saliency prediction, we develop a new DNN architecture that combines OM-CNN and 2C-LSTM together. Inspired by the second and third findings of Sections \ref{database_analysis}, OM-CNN integrates both regions and motions of objects to predict video saliency through two subnets, i.e., the subnets of objectness and motion. In OM-CNN, the objectness subnet yields a coarse objectness map, which is used to mask the features output from the \emph{convolutional layers} in the motion subnet. Then, the spatial features from the objectness subnet and temporal features from the motion subnet are concatenated to generate spatio-temporal features of OM-CNN. The architecture of OM-CNN is shown in Figure \ref{fOMCNN}.  According to the first finding of Section \ref{database_analysis}, 2C-LSTM with Bayesian dropout is developed to learn dynamic saliency of video clips, in which the spatio-temporal features of OM-CNN work as the input. Finally, the saliency map of each frame is generated from 2 \emph{deconvolutional layers} of 2C-LSTM. The architecture of 2C-LSTM is shown in Figure \ref{fLSTM}.

In the following, we present the detail of OM-CNN via introducing the objectness subnet in Section \ref{obsubnet} and the motion subnet in Section \ref{mosubnet}. In addition, the detail of 2C-LSTM is discussed in Section \ref{convlstm}. The training process of our DNN method is presented in Section \ref{cnntrain}.

%
%

\subsection{Objectness subnet in OM-CNN}\label{obsubnet}
The second finding of Section \ref{database_analysis} has shown that objects draw extensive attention in videos. Therefore, OM-CNN includes an objectness subnet, for extracting multi-scale spatial features related to objectness information. The basic structure of the objectness subnet is based on a pre-trained YOLO \cite{Redmon2015}. Note that YOLO is a state-of-the-art CNN architecture, capable of detecting video object with high accuracy. In OM-CNN, the YOLO structure is utilized to learn spatial features of the input frame for saliency prediction. To avoid over-fitting, the fast version of YOLO is applied in the objectness subnet, including 9 \emph{convolutional layers}, 5 \emph{pooling layers} and 2 \emph{Fully Connected layers} (\emph{FC}). To further avoid over-fitting, an additional \emph{batch-normalization layer} is added to each \emph{convolutional layer}. Assuming that $P(\cdot)$ and $\ast$ are the max pooling and convolution operations, the output of the $k$-th \emph{convolutional layer} $\mathbf{C}_o^k$ in the objectness subnet can be computed as
\begin{equation}\label{Convobj}
\mathbf{C}_o^k= L_{0.1}(P(\mathbf{C}_o^{k-1})\ast\mathbf{W}_o^{k-1}+\mathbf{B}_o^{k-1}),
\end{equation}
where $\mathbf{W}_o^{k-1}$ and $\mathbf{B}_o^{k-1}$ indicates the kernel parameters of weight and bias at the $(k-1)$-th \emph{convolutional layer}. Additionally, $L_{0.1}(\cdot)$ is a leaky ReLU activation with leakage coefficient of $0.1$.


For leveraging the multi-scale information with various receptive fields, \emph{feature normalization} (\emph{FN}) operation is introduced in OM-CNN to normalize and concatenate certain \emph{convolutional layers} of the objectness subnet. As shown in Figure \ref{fOMCNN}-(b), \emph{FN} includes a $1\times1$ \emph{convolutional layer} and a \emph{bilinear layer} to normalize the input features into 128 channels with size of $28\times28$. Specifically, in the objectness subnet, the outputs of the $4$-th, $5$-th, $6$-th and last \emph{convolutional layer} are normalized by \emph{FN} to obtain 4 sets of spatial features $\{\mathbf{FS}_i\}_{i=1}^4$, referring to multiple scales. Besides, the output of the last \emph{FC layer} in YOLO indicates the sizes, class probabilities and confidence of candidate objects in each grid. Then, the output of the last \emph{FC layer} needs to be reshaped into a spatial feature with size of $7\times7\times30$ (representing 30 channels with the size of $7\times7$). After bilinearly interpolating on the $7\times7\times30$ spatial feature, the high level spatial feature $\mathbf{FS}_5$ can be obtained with size of $28\times28\times30$. At last, the final spatial features are generated by concatenating $\{\mathbf{FS}_i\}_{i=1}^5$.

Given the spatial features $\{\mathbf{FS}_i\}_{i=1}^5$, an \emph{inference module} $I_c$ is designed to generate a coarse objectness map $\mathbf{S}_c$:
\begin{equation}\label{Coarseinfer}
\mathbf{S}_c=I_c(\{\mathbf{FS}_i\}_{i=1}^5).
\end{equation}
The \emph{inference module} $I_c$ is a CNN structure consisting of 4 \emph{convolutional layers} and 2 \emph{deconvolutional layers} with stride of $2$. The architecture of $I_c$ is shown in Figure \ref{fOMCNN}-(b). Consequently, the coarse objectness map $\mathbf{S}_c$ can be obtained to encode the objectness information, roughly related to salient regions.

\subsection{Motion subnet in OM-CNN}\label{mosubnet}
Next, a motion subnet, also shown in Figure \ref{fOMCNN}-(a), is incorporated in OM-CNN to extract multi-scale temporal features from the pair of neighboring frames. According to the third findings of Section \ref{database_analysis}, attention is more likely to be attracted by the moving objects or moving parts of the objects. Therefore, following the object subnet, the motion subnet is developed to extract motion features within object regions. The motion subnet is based on FlowNet \cite{dosovitskiy2015flownet}, a CNN structure to estimate optical flow. In the motion subnet, only the first 10 \emph{convolutional layers} of FlowNet are applied, in order to reduce the number of parameters. To model motion in object regions, the coarse objectness map $\mathbf{S}_c$ in \eqref{Coarseinfer} is used to mask the outputs of first 6 \emph{convolutional layers} of the motion subnet. As such, the output of the $k$-th \emph{convolutional layer} $\mathbf{C}_m^k$ can be computed as
\begin{eqnarray}\label{Convmotion}
&\mathbf{C}_m^k= L_{0.1}(M(\mathbf{C}_m^{k-1},\mathbf{S}_c)\ast\mathbf{W}_m^{k-1}+\mathbf{B}_m^{k-1}), \nonumber\\
\rm{where} &\quad M(\mathbf{C}_m^{k-1},\mathbf{S}_c)=\mathbf{C}_m^{k-1}\cdot(\mathbf{S}_c\cdot(1-\gamma)+\textbf{1}\cdot\gamma).
\end{eqnarray}
In \eqref{Convmotion}, $\mathbf{W}_m^{k-1}$ and $\mathbf{B}_m^{k-1}$ indicate the kernel parameters of weight and bias at the $(k-1)$-th \emph{convolutional layer} in the motion subnet; $\gamma$ ($0\leq\gamma\leq1$) is the adjustable parameter to control the mask degree, mapping the range of $\mathbf{S}_c$ from $[0,1]$ to $[\gamma,1]$. Note that the last 4 \emph{convolutional layers} are not masked with the coarse objectness map, for considering the motion of non-object region in saliency prediction. Afterwards, similar to the objectness subnet, the outputs of the $4$-th, $6$-th, $8$-th and $10$-th \emph{convolutional layers} are computed by FN, such that 4 sets of temporal features $\{\mathbf{FT}_i\}_{i=1}^4$ with size of $28\times28\times128$ are achieved.

Then, given the extracted features $\{\mathbf{FS}_i\}_{i=1}^5$ and $\{\mathbf{FT}_i\}_{i=1}^4$ from two subnets of OM-CNN, another \emph{inference module} $I_f$ is constructed to generate a fine saliency map $\mathbf{S}_f$, modeling the intra-frame saliency of a video frame. Mathematically, $\mathbf{S}_f$ can be computed as
\begin{equation}\label{Fineinfer}
\mathbf{S}_f=I_f(\{\mathbf{FS}_i\}_{i=1}^5,\{\mathbf{FT}_i\}_{i=1}^4).
\end{equation}
Here, $\mathbf{S}_f$ is also used to train the OM-CNN model, to be discussed in Section \ref{cnntrain}. Besides, the architecture of $I_f$ is same as that of $I_c$. As shown in Figure \ref{fOMCNN}-(b), in $I_f$, the output of \emph{convolutional layer} $C_4$ with size of $28\times28\times128$ is viewed as the final spatio-temporal features, denoted as $\mathbf{F}_{st}$, to predict intra-frame saliency. Next, $\mathbf{F}_{st}$ is fed into 2C-LSTM, to be presented in the following.

\subsection{Convolutional LSTM}\label{convlstm}
\begin{figure}
\begin{center}
\includegraphics[width=.9\linewidth]{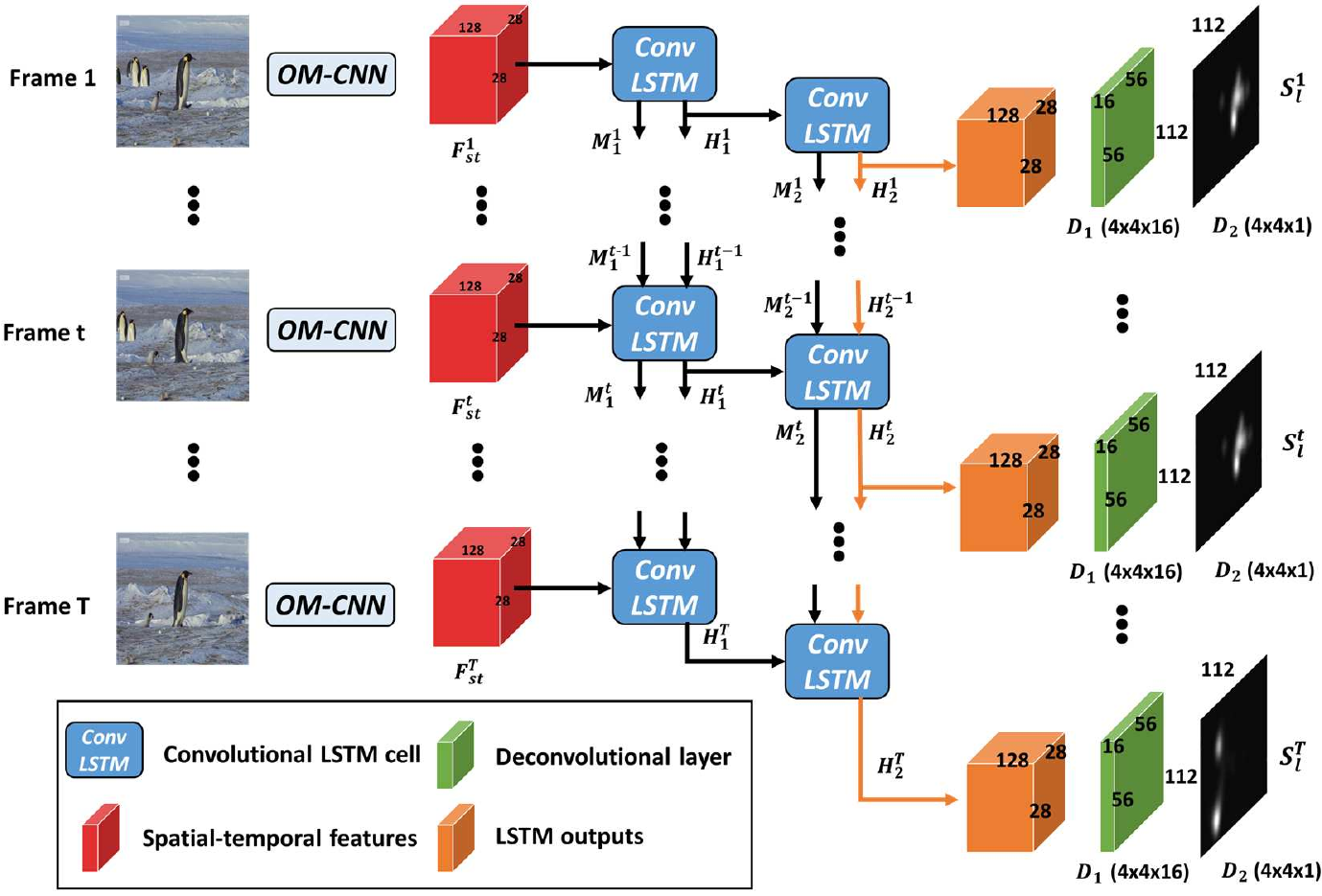}
\end{center}
\caption{\footnotesize \footnotesize Architecture of our 2C-LSTM for predicting saliency transition across inter-frame, following the OM-CNN. Note that the training process is not annotated in the figure.}\label{fLSTM}
\end{figure}

In this section, we develop the 2C-LSTM network for learning to predict dynamic saliency of a video clip, since the first finding of Section \ref{database_analysis} illustrates that there exists dynamic transition of attention across video frames. At frame $t$, taking the OM-CNN features $\mathbf{F}_{st}$ as the input (denoted as $\mathbf{F}_{st}^{t}$), 2C-LSTM leverages both long- and short-term correlation of the input features, through the memory cells ($\mathbf{M}_1^{t-1}, \mathbf{M}_2^{t-1}$) and hidden states ($\mathbf{H}_1^{t-1}, \mathbf{H}_2^{t-1}$) of 1-st and 2-nd LSTM layers at the last frame. Then, the hidden states of the 2-nd LSTM layer $\mathbf{H}_2^t$ are fed into 2 \emph{deconvolutional layers} to generate final saliency map $\mathbf{S}_l^{t}$ at frame t. The architecture of 2C-LSTM is shown in Figure \ref{fLSTM}.

The same as \cite{xingjian2015convolutional}, we extend the conventional LSTM via replacing Hadamard product (denoted as $\circ$) by the convolutional operator (denoted as $\ast$), in order to consider spatial correlation of input OM-CNN features in the dynamic model. Taking the first LSTM layer as example, a single LSTM cell at frame $t$ can be written as
\begin{eqnarray}\label{Convlstmeq}
 \mathbf{I}_1^{t} &=& \sigma(\mathbf{H}_1^{t-1}\ast\mathbf{W}_i^h+\mathbf{F}_{st}^{t}\ast\mathbf{W}_i^f+\mathbf{B}_i),\nonumber\\
 \mathbf{A}_1^{t} &=& \sigma(\mathbf{H}_1^{t-1}\ast\mathbf{W}_a^h+\mathbf{F}_{st}^{t}\ast\mathbf{W}_a^f+\mathbf{B}_a),\nonumber\\
 \mathbf{O}_1^{t} &=& \sigma(\mathbf{H}_1^{t-1}\ast\mathbf{W}_o^h+\mathbf{F}_{st}^{t}\ast\mathbf{W}_o^f+\mathbf{B}_o),\nonumber\\
 \mathbf{G}_1^{t} &=& \rm{tanh}(\mathbf{H}_1^{t-1}\ast\mathbf{W}_g^h+\mathbf{F}_{st}^{t}\ast\mathbf{W}_g^f+\mathbf{B}_g),\nonumber\\
 \mathbf{M}_1^{t} &=& \mathbf{F}_1^{t}\circ\mathbf{M}_1^{t-1}+\mathbf{I}_1^{t}\circ\mathbf{G}_1^{t},\nonumber\\
 \mathbf{H}_1^{t} &=& \mathbf{O}_1^{t}\circ\rm{tanh}(\mathbf{M}_1^{t}),
\end{eqnarray}
where $\sigma$ and $\rm{tanh}$ are the activation functions of sigmoid and hyperbolic tangent. In \eqref{Convlstmeq}, $\{\mathbf{W}_i^h, \mathbf{W}_a^h, \mathbf{W}_o^h, \mathbf{W}_g^h,\mathbf{W}_i^f, \mathbf{W}_a^f, \mathbf{W}_o^f, \mathbf{W}_g^f\}$  and $\{\mathbf{B}_i, \mathbf{B}_a, \mathbf{B}_o, \mathbf{B}_g\}$  denote the kernel parameters of weight and bias at the corresponding \emph{convolutional layer}; $\mathbf{I}_1^{t}$, $\mathbf{A}_1^{t}$ and $\mathbf{O}_1^{t}$ are the the gates of input, forget and output for frame $t$; $\mathbf{G}_1^{t}$, $\mathbf{M}_1^{t}$ and $\mathbf{H}_1^{t}$ are the input modulation, memory cells and hidden states. They are all represented by 3-D tensors with size of $28\times28\times128$ in 2C-LSTM.

In our method, we adopt two-layer LSTM cells to learn temporal correlation of high-dimensional features ($28\times28\times128$). On the other hand, the two-layer LSTM cells decrease the generalization ability. Thus, to improve generalization ability, we apply the Bayesian inference based dropout \cite{gal2016theoretically} in each convolutional LSTM cell. Then, the LSTM cell of \eqref{Convlstmeq} can be rewritten as
\begin{eqnarray}\label{Convlstmeq_dp}
 \mathbf{I}_1^{t} &=& \sigma((\mathbf{H}_1^{t-1}\circ\mathbf{Z}_i^h)\ast\mathbf{W}_i^h+(\mathbf{F}_{st}^{t}\circ\mathbf{Z}_i^f)\ast\mathbf{W}_i^f+\mathbf{B}_i),\nonumber\\
 \mathbf{A}_1^{t} &=& \sigma((\mathbf{H}_1^{t-1}\circ\mathbf{Z}_a^h)\ast\mathbf{W}_a^h+(\mathbf{F}_{st}^{t}\circ\mathbf{Z}_a^f)\ast\mathbf{W}_a^f+\mathbf{B}_a),\nonumber\\
 \mathbf{O}_1^{t} &=& \sigma((\mathbf{H}_1^{t-1}\circ\mathbf{Z}_o^h)\ast\mathbf{W}_o^h+(\mathbf{F}_{st}^{t}\circ\mathbf{Z}_o^f)\ast\mathbf{W}_o^f+\mathbf{B}_o),\nonumber\\
 \mathbf{G}_1^{t} &=& \rm{tanh}((\mathbf{H}_1^{t-1}\circ\mathbf{Z}_g^h)\ast\mathbf{W}_g^h+(\mathbf{F}_{st}^{t}\circ\mathbf{Z}_g^f)\ast\mathbf{W}_g^f+\mathbf{B}_g),\nonumber\\
 \mathbf{M}_1^{t} &=& \mathbf{F}_1^{t}\circ\mathbf{M}_1^{t-1} + \mathbf{I}_1^{t}\circ\mathbf{G}_1^{t},\nonumber\\
 \mathbf{H}_1^{t} &=& \mathbf{O}_1^{t}\circ\rm{tanh}(\mathbf{M}_1^{t}).
\end{eqnarray}
In \eqref{Convlstmeq_dp}, $\{\mathbf{Z}_i^h, \mathbf{Z}_a^h, \mathbf{Z}_o^h, \mathbf{Z}_h^h\}$ and $\{\mathbf{Z}_i^f, \mathbf{Z}_a^f, \mathbf{Z}_o^f, \mathbf{Z}_g^f\}$  are 2 sets of random masks for the hidden states and input features before convolutional operation. These masks are generated by a $L$-times Monte Carlo integration, with hidden dropout rate $p_h$ and feature dropout rate $p_f$, respectively.

\subsection{Training process}\label{cnntrain}
For training OM-CNN, we utilize the Kullback-Leibler (KL) divergence based loss function to update the parameters. It is because \cite{huang2015salicon} has proved that the KL divergence is more effective than other metrics in training DNN to predict saliency. Regarding the saliency map as a probability distribution of attention, we can measure the KL divergence $D_{\rm{KL}}$ between fine saliency map $\mathbf{S}_f$ of OM-CNN and ground truth distribution $\mathbf{G}$ of human fixations as follows,
\begin{equation}\label{KLdiv}
D_{\rm{KL}}(\mathbf{G},\mathbf{S}_f)=\mathbf{G}\log\frac{\mathbf{G}}{\mathbf{S}_f}.
\end{equation}
In \eqref{KLdiv}, the smaller KL divergence indicates higher accuracy in saliency prediction. Furthermore, the KL divergence between coarse objectness map $\mathbf{S}_c$ of OM-CNN and ground truth $\mathbf{G}$ is also used as an auxiliary function to train OM-CNN. This is based on the assumption that the object regions are correlated with salient regions. Then, the OM-CNN model is trained by minimizing the following loss function,
\begin{equation}\label{OMCNNloss}
L_{\rm{OM-CNN}} = \frac{1}{1+\lambda}D_{\rm{KL}}(\mathbf{G},\mathbf{S}_f) + \frac{\lambda}{1+\lambda}D_{\rm{KL}}(\mathbf{G},\mathbf{S}_c).
\end{equation}
In \eqref{OMCNNloss}, $\lambda$ is a hyper-parameter to control the weights of two KL divergences. Note that OM-CNN is pre-trained on YOLO and FlowNet, and the rest parameters of OM-CNN are initialized by the Xavier initializer \cite{glorot2010understanding}.

To train 2C-LSTM, the training videos are cut into clips with the same length $T$. Besides, when training 2C-LSTM, the parameters of OM-CNN are fixed to extract the spatio-temporal features of each $T$-frame video clip. Then, the loss function of 2C-LSTM is defined as the averaged KL divergence over $T$ frames
\begin{equation}\label{lstmloss}
L_{\rm{2C-LSTM}} = \frac{1}{T}\sum_{i=1}^{T}D_{\rm{KL}}(\mathbf{S}_l^i,\mathbf{G}_i).
\end{equation}
In \eqref{lstmloss}, $\{\mathbf{S}_l^i\}_{i=1}^{T}$ are the final saliency maps generated by 2C-LSTM, and $\{\mathbf{G}_i\}_{i=1}^{T}$ are the ground truth of attention maps. For each LSTM cell, the kernel parameters are initialized by the Xavier initializer, while the memory cells and hidden states are initialized by zeros.


\section{Experiment}\label{experiment}
In this section, the experimental results are presented to validate the performance of our method in video saliency prediction. Section \ref{setting} introduces the settings in our experiments. Section \ref{evaour} and \ref{evaother} compare the accuracy of saliency prediction on LEDOV and other 2 public databases, respectively. Furthermore, the results of ablation experiment are discussed in Section \ref{ablation}, to analyze the effectiveness of each individual component proposed in our method.
\subsection{Settings}\label{setting}
In our experiment, 538 videos in the LEDOV database are randomly divided into training (436 videos), validation (41 videos) and test (41 videos) sets. Specifically, to learn 2C-LSTM, we temporally segment 456 training videos into 24,685 clips, all of which have $T=16$ frames. The overlap of 10 frames is allowed in cutting video clips, for the purpose of data augmentation.
Before inputting to OM-CNN, the RGB channels of each frame are resized to $448\times448$, with their mean values being removed. In training OM-CNN and 2C-LSTM, we learn the parameters using the stochastic gradient descent algorithm with Adam optimizer \cite{kingma2014adam}. Here, the hyper-parameters of OM-CNN and 2C-LSTM are tuned to minimize the KL divergence of saliency prediction over the validation set. The tuned values of some key hyper-parameters are listed in Table \ref{Parameters}. Given the trained models of OM-CNN and 2C-LSTM, all 41 test videos in LEDOV are used to evaluate the performance of our method, compared with other 8 state-of-the-art methods. All experiments are conducted on a computer with Intel(R) Core(TM) i7-4770 CPU@3.4 GHz, 16 GB RAM and a single GPU of Nvidia GeForce GTX 1080. Benefitting from the acceleration of GPU, our method is able to make the real-time prediction for video saliency at a speed of 30 fps.

\begin{table}
\vspace{-2em}
\tiny
\caption{ \footnotesize{The values of hyper-parameters in OM-CNN and 2C-LSTM.}}\label{Parameters}
\begin{center}
\begin{tabular}{|c| l|l|}
 \hline
  \multirow{6}*{Hyper-parameters in OM-CNN}&Objectness mask parameter $\gamma$ in \eqref{Convmotion}      &  0.5 \\
  &KL divergences weight $\lambda$ in \eqref{OMCNNloss}    &  0.5 \\
  &Initial learning rate     &  $1\times10^{-5}$\\
  &Training epochs (iterations)  & $12(\sim1.5\times10^5)$\\
  &Batch size  & 12\\
  &Weight decay  & $5\times10^{-6}$\\
  \hline
  \multirow{5}*{Hyper-parameters in 2C-LSTM}& Bayesian dropout rates $p_h$ and $p_f$ in 2C-LSTM    &  0.25\&0.25\\
  &Times of Monte Carlo integration $L$ in in 2C-LSTM  & 100\\
  &Initial learning rate   &  $1\times10^{-4}$\\
  &Training epochs (iterations)  & $15(\sim2\times10^5)$\\
  &Weight decay  & $5\times10^{-6}$\\
   \hline
\end{tabular}
\end{center}
\end{table}

\begin{table*}[t]
\footnotesize
  \centering
  \caption{ Mean (standard deviation) of saliency prediction accuracy for our and other 8 methods over all test videos in LEDOV.}
  \vspace{-1em}
 \begin{tabular}{cccccccccc}
    \toprule
          & Our   & GBVS \cite{harel2006graph}  & PQFT \cite{guo2010novel} &  Rudoy \cite{rudoy2013iccv} & OBDL \cite{Khatoonabadi2015CVPR} & SALICON \cite{huang2015salicon} & Xu \cite{xu2017learning}& BMS \cite{zhang2016exploiting} & SalGAN \cite{Pan_2017_SalGAN}\\
    \midrule
    AUC   & \textbf{0.90}(0.04) & 0.84(0.06) & 0.70(0.08) & 0.80(0.08) & 0.80(0.09) & 0.85(0.06) & 0.83(0.06) & 0.76(0.09) & 0.87(0.06) \\
    NSS   & \textbf{2.94}(0.85) & 1.54(0.74) & 0.69(0.46) & 1.45(0.64) & 1.54(0.84) & 2.33(0.87) & 1.47(0.47) & 0.98(0.48) & 2.19(0.59) \\
    CC    & \textbf{0.57}(0.12) & 0.32(0.13) & 0.14(0.08) & 0.32(0.14) & 0.32(0.16) & 0.44(0.13) & 0.38(0.11) & 0.21(0.09) & 0.43(0.09) \\
    KL    & \textbf{1.24}(0.39) & 1.82(0.39) & 2.46(0.39) & 2.42(1.53) & 2.05(0.74) & 1.64(0.42) & 1.65(0.30) & 2.23(0.39) & 1.68(0.33) \\
    \bottomrule
    \end{tabular}%
  \label{saliency_accuracy}%
  \vspace{-1em}
\end{table*}%

\begin{figure*}[t]
\begin{center}
\includegraphics[width=.9\linewidth]{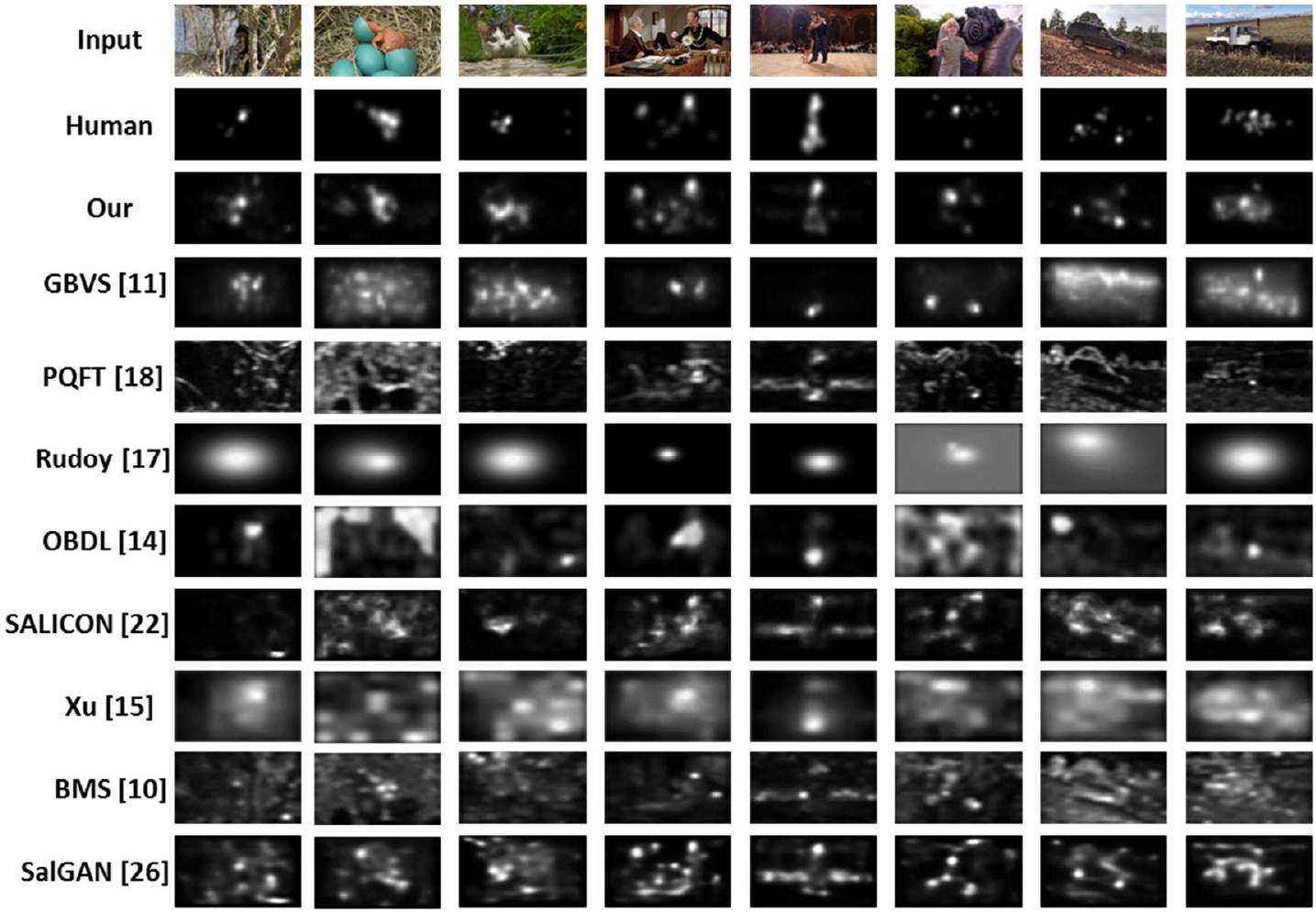}
  \end{center}
\caption{\footnotesize Saliency maps of 8 videos randomly selected from the test set of LEDOV. The maps were yielded by our and other 8 methods as well the ground-truth human fixations. Note that the results of only one frame are shown for each selected video.}\label{subfigure}
\end{figure*}

\subsection{Evaluation on LEDOV}\label{evaour}
In this section, we compare the accuracy of video saliency predicted by our and other 8 state-of-the-art methods, including GBVS \cite{harel2006graph}, PQFT \cite{guo2010novel}, Rudoy \cite{rudoy2013iccv}, OBDL \cite{Khatoonabadi2015CVPR}, SALICON \cite{huang2015salicon}, Xu \cite{xu2017learning}, BMS \cite{zhang2016exploiting} and SalGAN \cite{Pan_2017_SalGAN}. Among them, \cite{harel2006graph}, \cite{guo2010novel}, \cite{rudoy2013iccv}, \cite{Khatoonabadi2015CVPR} and \cite{xu2017learning} are 5 state-of-the-art saliency prediction methods for videos. Besides, we compare two DNN based methods, \cite{huang2015salicon} and \cite{Pan_2017_SalGAN}. In our experiments, we apply four metrics to measure the accuracy of saliency prediction: the area under receiver operating characteristic curve (AUC), normalized scanpath saliency (NSS), CC, and KL divergence. Note that the larger value of AUC, NSS or CC indicates more accurate prediction of saliency, and the smaller KL divergence means better saliency prediction. Table \ref{saliency_accuracy} tabulates the results of AUC, NSS, CC and KL divergence for our and other 8 methods, which are averaged over 41 test videos of the LEDOV database. We can see from this table that our method performs much better than all other methods in all 4 metrics. More specifically, our method achieves at least 0.03, 0.61, 0.13 and 0.40 improvements in AUC, NSS, CC and KL. Besides, two DNN based methods, SALICON \cite{huang2015salicon} and SalGAN \cite{Pan_2017_SalGAN}, outperform other conventional methods. This verifies the effectiveness of saliency related features that are automatically learned by DNN, better than hand crafted features. On the other hand, our method is significantly superior to \cite{huang2015salicon} and \cite{Pan_2017_SalGAN}. The main reasons are as follows: (1) Our method embeds the objectness subnet to make use of objectness information in saliency prediction; (2) The object motion is explored in the motion subnet to predict video saliency. (3)The network of 2C-LSTM is leveraged to model saliency transition across video frames.
Section \ref{ablation} analyzes the above three reasons with more details.

\begin{figure}
\begin{center}
\includegraphics[width=.99\linewidth]{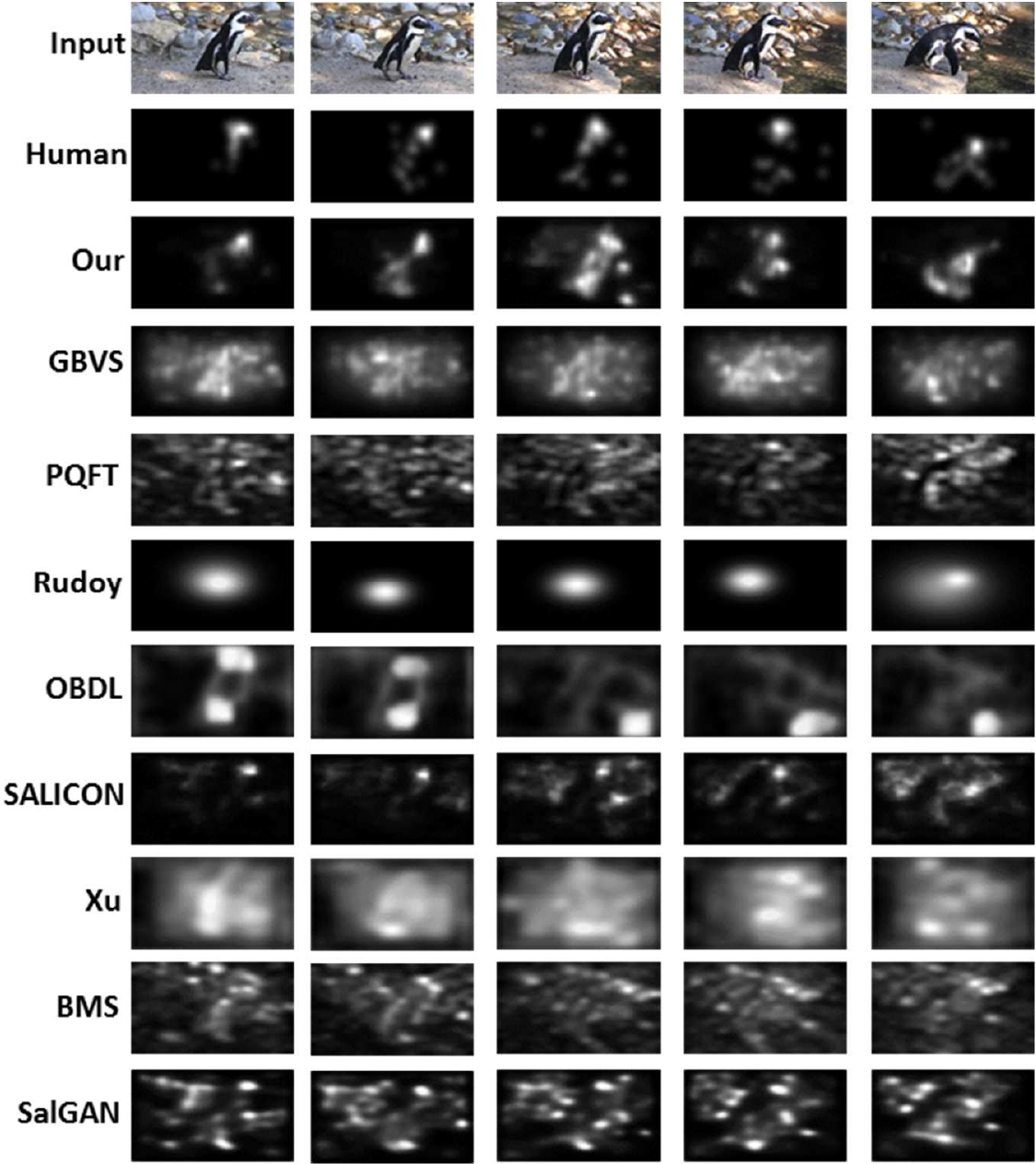}
  \end{center}
\caption{\footnotesize Saliency maps of several frames randomly selected from a single test video in LEDOV. The maps were yielded by our and other 8 methods as well the ground-truth human fixations.}\label{subfigure2}
\end{figure}

Next, we move to the comparison of subjective results in video saliency prediction. Figure \ref{subfigure} demonstrates the saliency maps of 8 randomly selected videos in test set, detected by our and other 8 methods. In this figure, one frame is selected for each video. One may observe from Figure \ref{subfigure} that our method is capable of well locating the salient regions, much closer to the ground-truth maps of human fixations. In contrast, most of other methods fail to accurately predict the regions attracting human attention. In addition, Figure \ref{subfigure2} shows the saliency maps of some frames selected from one test video. As seen in this figure, our method is able to model human fixation with smooth transition than other 8 methods. In summary, our method is superior to other state-of-the-art methods in both objective and subjective results, tested on our LEDOV database.

\subsection{Evaluation on other databases}\label{evaother}

To evaluate the generalization capability of our method, we further compare the performance of our and other 8 methods on two widely used databases SFU \cite{Hadizadeh2012SFU} and DIEM \cite{Mital2011DIEM}, which are available online. The same as \cite{Khatoonabadi2015CVPR}, 20 videos of DIEM and all videos of SFU are tested to assess the saliency prediction performance. In our experiments, the models of OMM-CNN and 2C-LSTM, learned from the training set of LEDOV, are directly used to predict saliency of test videos from the DIEM and SFU databases. Table \ref{evaluation_other} presents the averaged results of AUC, NSS, CC and KL for our and other 8 methods over SFU and DIEM, respectively. We can see from this table that our method again outperforms other 8 methods, especially in the DIEM database. In particular, there are at least 0.04, 0.44, 0.09 and 0.25 improvements in AUC, NSS, CC and KL. Such improvements are comparable to those in our LEDOV database. This implies the generalization capability of our method in video saliency prediction.

\begin{table*}
\footnotesize
  \centering
  \caption{Mean (standard deviation) values for saliency prediction accuracy of our and other methods over SFU and DIEM databases.}
  \vspace{-1em}
    \begin{tabular}{|c|ccccccccc|}

     \hline
          &\multicolumn{9} {|c|} {SFU}\\
    \cline{2-10}
          & Our   & GBVS \cite{harel2006graph}  & PQFT \cite{guo2010novel} &  Rudoy \cite{rudoy2013iccv} & OBDL \cite{Khatoonabadi2015CVPR} & SALICON \cite{huang2015salicon} & Xu \cite{xu2017learning}& BMS \cite{zhang2016exploiting} & SalGAN \cite{Pan_2017_SalGAN}\\
        \hline
    AUC   & \textbf{0.81}(0.07) & 0.76(0.07) & 0.61(0.09) & 0.73(0.08) & 0.74(0.10) & 0.78(0.08) & 0.80(0.07) & 0.66(0.08) & 0.79(0.07) \\
    NSS   & \textbf{1.46}(0.65) & 0.91(0.47) & 0.31(0.34) & 0.83(0.45) & 1.03(0.64) & 1.24(0.60) & 1.24(0.39) & 0.50(0.31) & 1.25(0.47) \\
    CC    & \textbf{0.55}(0.15) & 0.44(0.15) & 0.12(0.15) & 0.34(0.15) & 0.42(0.21) & 0.58(0.22) & 0.43(0.12) & 0.25(0.11) & 0.51(0.13) \\
    KL    & 0.67(0.24) & \textbf{0.61}(0.19) & 0.98(0.27) & 0.93(0.36) & 0.80(0.33) & 1.12(1.76) & 1.35(0.25) & 0.83(0.20) & 0.70(0.25) \\
    \hline
        &\multicolumn{9} {|c|} {DIEM}\\
    \cline{2-10}
          & Our   & GBVS \cite{harel2006graph}  & PQFT \cite{guo2010novel} &  Rudoy \cite{rudoy2013iccv} & OBDL \cite{Khatoonabadi2015CVPR} & SALICON \cite{huang2015salicon} & Xu \cite{xu2017learning}& BMS \cite{zhang2016exploiting} & SalGAN \cite{Pan_2017_SalGAN}\\
        \hline
    AUC   & \textbf{0.86}(0.08) & 0.81(0.09) & 0.71(0.11) & 0.80(0.11) & 0.75(0.14) & 0.79(0.11) & 0.80(0.11) & 0.77(0.11) & 0.81(0.08) \\
    NSS   & \textbf{2.25}(1.16) & 1.21(0.82) & 0.86(0.71) & 1.40(0.83) & 1.26(1.03) & 1.68(1.04) & 1.34(0.74) & 1.20(0.80) & 1.60(0.71) \\
    CC    & \textbf{0.49}(0.21) & 0.30(0.18) & 0.19(0.14) & 0.38(0.20) & 0.29(0.22) & 0.36(0.19) & 0.35(0.17) & 0.28(0.17) & 0.35(0.13) \\
    KL    & \textbf{1.30}(0.55) & 1.64(0.48) & 1.73(0.44) & 2.33(2.05) & 2.77(1.58) & 1.66(0.58) & 1.67(0.39) & 1.96(1.13) & 1.64(0.41) \\
    \hline
    \end{tabular}%
  \label{evaluation_other}%
\end{table*}%

\subsection{Performance analysis of saliency prediction}\label{ablation}
\begin{figure}
\begin{center}
\includegraphics[width=.9\linewidth]{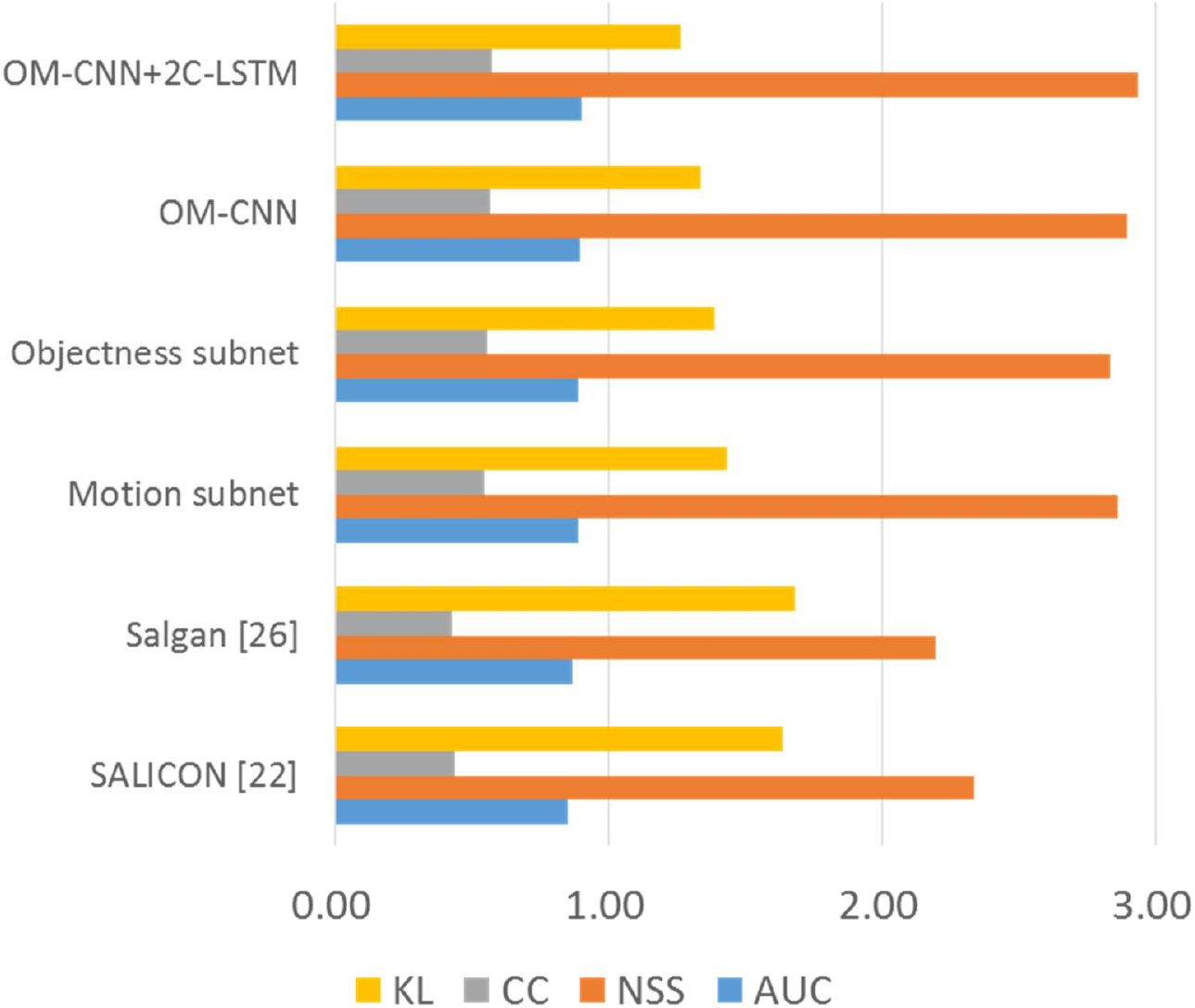}
  \end{center}
\caption{\footnotesize Saliency prediction accuracy of objectness subnet, motion subnet, OM-CNN and the combination of OM-CNN and 2C-LSTM, compared with SALICON \cite{huang2015salicon} and SalGAN \cite{Pan_2017_SalGAN}. Note that the smaller KL divergence indicates higher accuracy in saliency prediction.}\label{ablationfigure}
\end{figure}

Since the OM-CNN architecture of our method is composed of the objectness and motion subnets, we evaluate the contribution of each single subnet. We further analyze the contribution of 2C-LSTM, by comparing the trained models with and without 2C-LSTM. Specifically, the objectness subnet, motion subnet and OM-CNN are trained independently, with the same settings introduced above. Then, they are compared with our method, i.e., the combination of OM-CNN and 2C-LSTM. The comparison results are shown in Figure \ref{ablationfigure}. We can see from this figure that OM-CNN performs better than the objectness subnet with 0.05 reduction in KL divergence, and outperforms the motion subnet with 0.09 KL divergence reduction. This indicates the effectiveness of integrating the subnets of objectness and motion. Besides, the combination of OM-CNN and 2C-LSTM reduces the KL divergence by 0.09, over the single OM-CNN architecture. Hence, we can conclude that 2C-LSTM can further improve the performance of OM-CNN, due to exploring temporal correlation of saliency across video frames.

\begin{figure}
\begin{center}
\includegraphics[width=.9\linewidth]{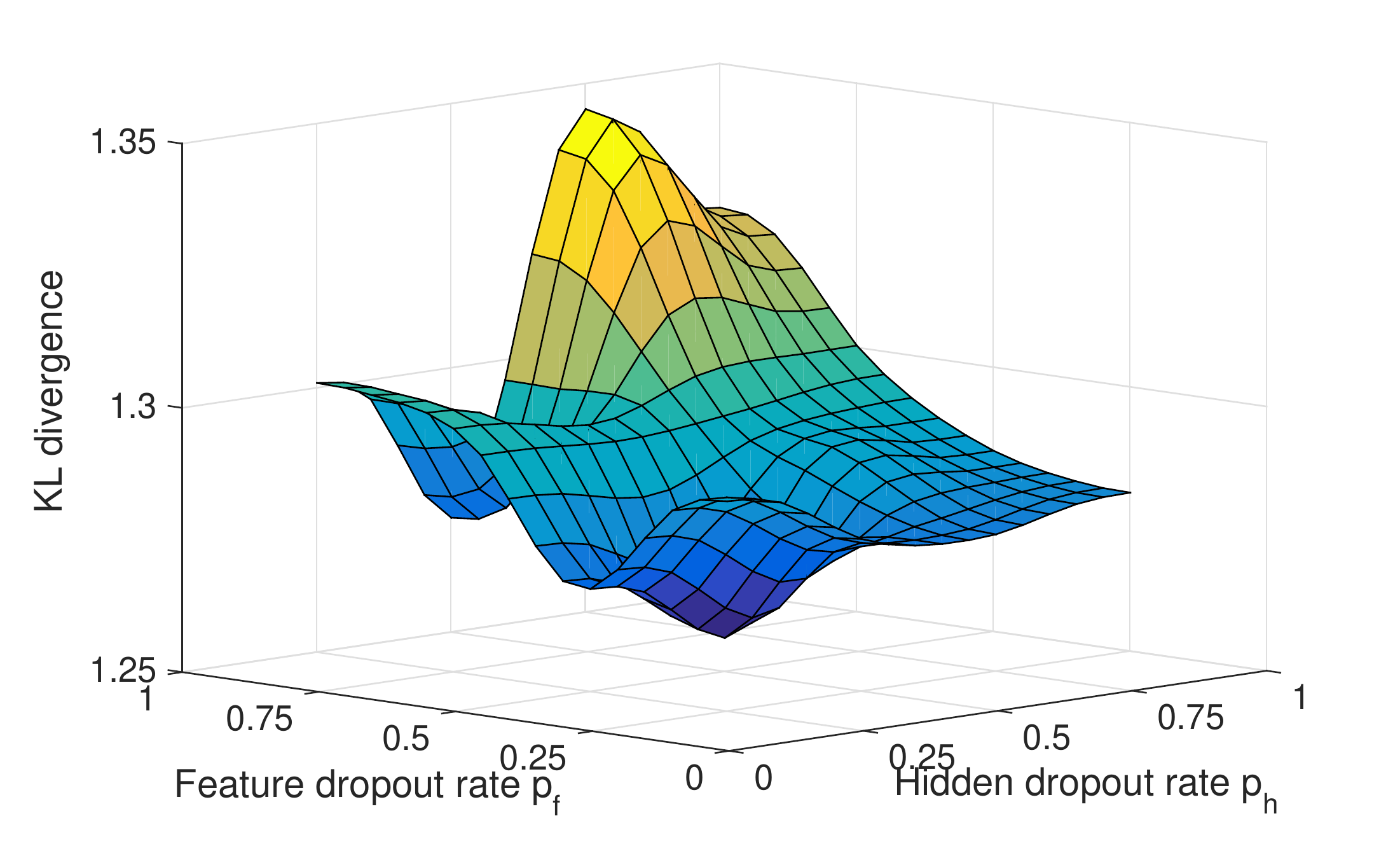}
  \end{center}
\caption{\footnotesize The KL divergences of our models, using different drop out rates in 2C-LSTM.}\label{dpratesfigure}
\end{figure}

Furthermore, we analyze the performance of Bayesian dropout in 2C-LSTM, which aims to avoid the over-fitting caused by the high dimensionality of 2C-LSTM. Through the experimental results, we find that the low dropout rate may incur the under-fitting issue, resulting in accuracy reduction of saliency prediction. To analyze the impact of the dropout rate, we train the 2C-LSTM models at different values of hidden dropout rate $p_h$ and feature dropout rate $p_f$.
The trained models are tested over the validation set of LEDOV, and the averaged results of KL divergence are shown in Figure \ref{dpratesfigure}. We can observe from the figure that the Bayesian dropout can bring around 0.03 KL reduction when both $p_h$ and $p_f$ are set to 0.25. However,
the KL divergence sharply rises, once $p_h$ and $p_f$ are increased from 0.25 to 1. Therefore, we set $p_h$ and $p_f$ to be 0.25 in our method. They may be adjusted according to the amount of training data.

\section{Conclusion}\label{conclusion}
In this paper, we have proposed a DNN based video saliency prediction method. In our method, two DNN architectures were developed, i.e., OM-CNN and 2C-LSTM. The two DNN architectures were driven by the LEDOV database established in this paper, which is composed of 32 subjects’ fixations on 538 videos. Interestingly, we found from the LEDOV database that the human fixations more likely fall into the objects, especially the moving objects or moving regions in the objects. Additionally, we found that the correlation of attention across consecutive frames is high. In light of these findings, the OM-CNN architecture was proposed to explore the spatio-temporal features of object and motion to predict the intra-frame saliency of videos, and the 2C-LSTM architecture was developed to model the inter-frame saliency of videos. Finally, the experimental results verified that our DNN based method significantly outperforms other 8 state-of-the-art methods over both our and other two public video eye-tracking databases, in terms of AUC, CC, NSS, and KL metrics.

There are two promising directions for future work. First, our method mainly focuses on videos with objects. Actually, a handful of videos are of natural scenes, without any salient object. Hence, saliency prediction of natural scene videos is an interesting research direction in future. The second future work is the potential application of our method in perceptual video coding. In particular, our method is able to locate salient and non-salient regions in videos, and it is expected that the coding efficiency of videos can be improved by removing the perceptual redundancy existing in the non-salient regions. Consequently, the less bits are needed to encode and deliver videos, greatly relieving the bandwidth-hungry issue in video transmission.

\footnotesize
\bibliographystyle{elsarticle-num}
\bibliography{ref}

\end{document}